\documentclass[journal]{IEEEtran}

\usepackage{cite}
\usepackage{graphicx}
\usepackage{hyperref}
\usepackage{amssymb}
\usepackage{amsmath}
\usepackage{pifont}
\usepackage{amsfonts}
\usepackage{subcaption}
\usepackage{booktabs}
\usepackage{algorithm}
\usepackage{algorithmic}
\usepackage{mathtools}
\usepackage{hyperref}
\makeatletter
\newcommand*{\rom}[1]{\expandafter\@slowromancap\romannumeral #1@}
\makeatother

\usepackage{tikz}
\newcommand*\circled[1]{\tikz[baseline=(char.base)]{
    \node[shape=circle,draw,inner sep=0.9pt] (char) {#1};}}

\usepackage[utf8]{inputenc}
\usepackage[english]{babel}
\usepackage[font=small,labelfont=bf]{caption}

\usepackage{amsthm}
\usepackage{xspace} 
 \usepackage{xcolor}
\definecolor{todocolor}{rgb}{0.9,0.1,0.1}
\definecolor{hycolor}{rgb}{0.7,0.7,0.3}

\theoremstyle{definition}
\newtheorem{theorem}{Theorem}

\theoremstyle{definition}
\newtheorem{definition}{Definition}

\newtheorem{lemma}{Lemma}
\newtheorem{corollary}{Corollary}
\newtheorem{proposition}{Proposition}

\theoremstyle{remark}
\newtheorem*{remark}{Remark}

\theoremstyle{definition}

\newcommand{\DPP}{\textsc{Dpp}\xspace}
\newcommand{\DML}{\textsc{Dml}\xspace}

\ifCLASSINFOpdf
\else
\fi
\begin{document}
%
\title{Secure Metric Learning via Differential Pairwise Privacy}
%
%
%

\author{Jing~Li, Yuangang~Pan, Yulei~Sui, and Ivor~W.~Tsang
\thanks{J. Li, Y. Pan, Y, Sui, and I. W. Tsang are with the Centre for Artificial Intelligence (CAI), University of Technology Sydney, Australia.}

}

\maketitle

\begin{abstract}
Distance Metric Learning (\DML) has drawn much attention over the last two decades. A number of previous works have shown that it performs well in measuring the similarities of individuals given a set of correctly labeled pairwise data by domain experts. These important and precisely-labeled pairwise data are often highly sensitive in real world (e.g., patients similarity). This paper studies, for the first time, how pairwise information can be leaked to attackers during distance metric learning, and develops differential pairwise privacy (\DPP), generalizing the definition of standard differential privacy, for secure metric learning.

Unlike traditional differential privacy which only applies to independent samples, thus cannot be used for pairwise data, \DPP successfully deals with this problem by reformulating \emph{the worst case}. Specifically, given the pairwise data, we reveal all the involved correlations among pairs in the constructed undirected graph. \DPP is then formalized that defines what kind of \DML algorithm is private to preserve pairwise data. After that, a case study employing the contrastive loss is exhibited to clarify the details of implementing a \DPP-\DML algorithm. Particularly, the sensitivity reduction technique is proposed to enhance the utility of the output distance metric. Experiments both on a toy dataset and benchmarks demonstrate that the proposed scheme achieves pairwise data privacy without compromising the output performance much (Accuracy declines less than 0.01 throughout all benchmark datasets when the privacy budget is set at 4).
\end{abstract}


%
\IEEEpeerreviewmaketitle

\section{Introduction}
\IEEEPARstart{T}{he} distance/similarity between two samples (e.g., euclidean distance) is the base of many applications \cite{bellet2013survey}, such as clustering, classification, information retrieval, etc. Distance Metric Learning (\DML) is a fundamental tool that learns a distance metric over the data to support these applications. It expects that in the projected space the similar samples would be better grouped; while the dissimilar ones would be appropriately separated. 
Such a principle is properly spoken of in \cite{xing2003distance} and much subsequent research \cite{tsang2003DML,tsang2003ideal,weinberger2009distance,guillaumin2009you,ying2012distance,hu2014discriminative,sohn2016improved,xie2018nonoverlap} has followed this criterion. 

\begin{figure}
 \centering
  \includegraphics[width=8.5cm]{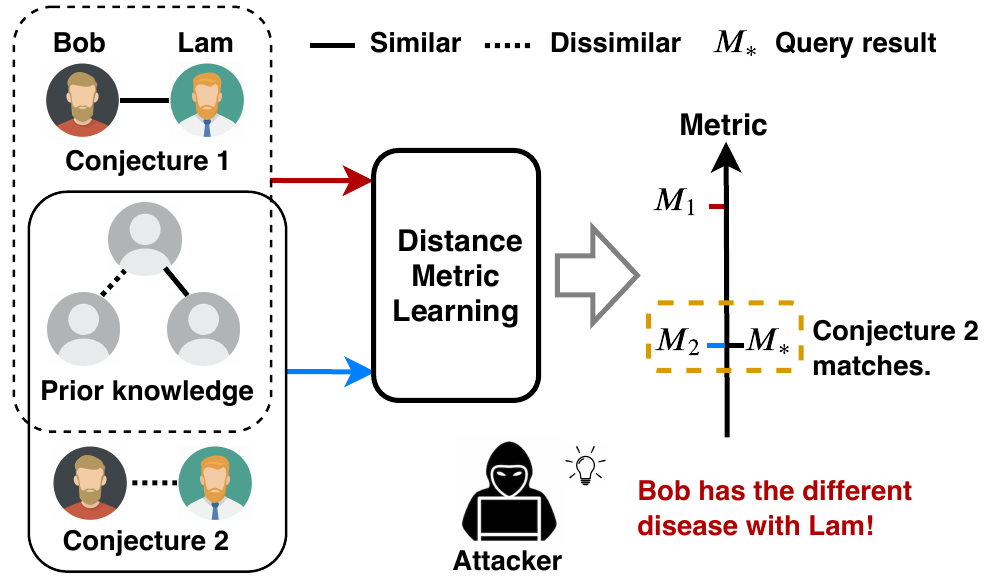}
  \caption{Leakage of pairwise relationship. An attacker with all the prior knowledge of the dataset except the target relationship between Bob and Lam, is able to infer their real relationship by matching the conjecture and query results. }
  \label{fig_why_we_do}
\end{figure}

\begin{figure*}
 \centering
  \includegraphics[width=15cm]{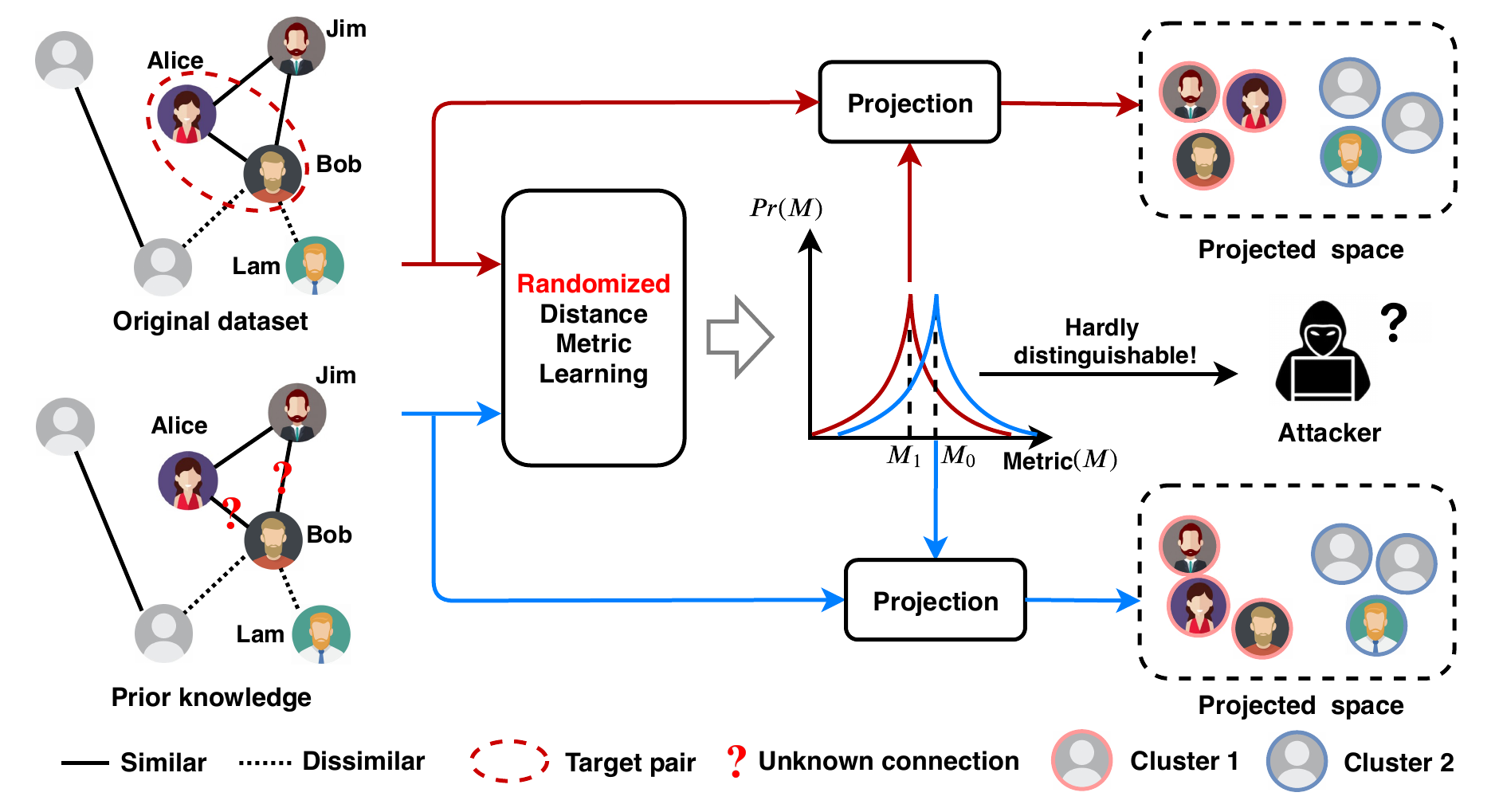}
  \caption{Preserving privacy of pairwise relationship. Suppose the relationship between Alice and Bob is the target. The attacker may have the prior knowledge that excludes edges with question mark. This provides one of the worst cases, where the relationship of Alice and Bob cannot be inferred from prior knowledge. \DPP ensures that the prior knowledge of the attacker for the worst case has the hardly indistinguishable output with the original dataset. Particularly, the obtained metric $M_0$ is expected to group training data as $M_1$ does. }
  \label{fig_how_we_do}
\end{figure*}

The training data fed to \DML model is often pairwise labelled\footnote{This setting is called weakly supervised \DML in some literature but is overwhelming in metric learning community.} which naturally encodes some secrets when the data is collected from humans. 
A popular application of learning distance metric would be healthcare data~\cite{wang2011integrating,huai2018uncorrelated,suo2019metric}.
A pairwise relationship in this application might be, for example, two patients have the same/different disease(s). However, the relationships in the training data can be leaked to attackers through a deterministic and resilient \DML model. We give a white-box attack example to explain a scenario that a traditional \DML model may unfortunately leak the pairwise relationship to external attackers.

\emph{\textbf{A privacy leakage scenario}.} Assume that we have a set of pairwise medical records as the training data, a \DML model is trained over the entire pairwise data and returns a static metric $M_*$ (e.g., oracle). Suppose that a powerful attacker understands the \DML model and has some prior knowledge, i.e., knowing partial pairwise data of the entire training set (e.g. $K-1$, where $K$ is the total number of pairs). The attacker would like to exploit a particular relationship (e.g.,  Bob and Lam) which the attacker does not know. First, she/he combines the prior knowledge by checking two possible conjectures: (1) Bob and Lam have the same disease, and (2) Bob and Lam have different diseases. Then, the attacker feeds them to the \DML model separately and obtains corresponding conjecture results $M_1$ and $M_2$. Lastly, she/he matches the two results with the oracle $M_*$. The matched conjecture will expose the correct pairwise relationship between Bob and Lam, leaking the private relationship to the attacker. Fig. \ref{fig_why_we_do} depicts this process.

\emph{\textbf{Differential privacy}.} 
\begin{figure}
 \centering
  \includegraphics[width=8.5cm]{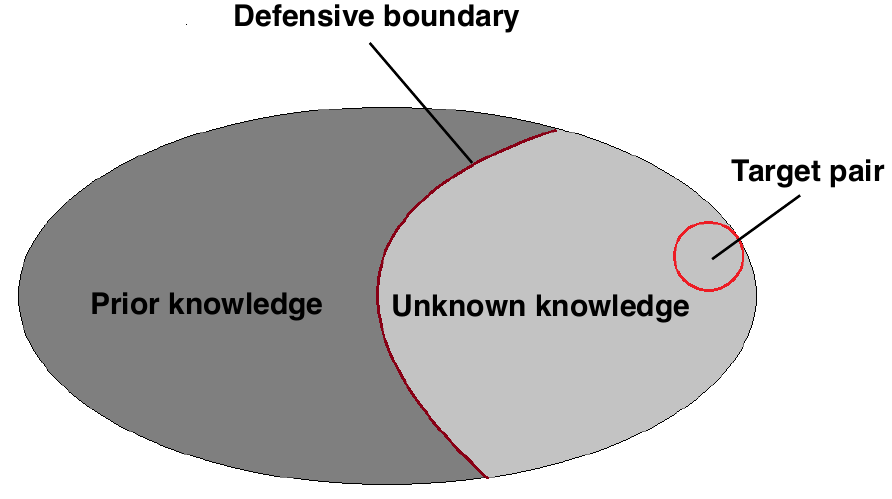}
  \caption{ Knowledge diagram. The prior knowledge is supposed to be smaller than the whole data deducting the target pair because of the data correlation. For a given target pair there always exists a corresponding defensive boundary which restricts the volume of prior knowledge in practice. }
  \label{boundary}
\end{figure}
Differential Privacy (DP) \cite{dwork2006calibrating} has become a popular and widely accepted privacy framework in recent years. It is initially used in data mining and data release research community and recently moved to the machine learning community\cite{ji2014differential}. It aims to ensure that the participation of an individual sample never changes the probability of any possible outcome by much. This goal is usually achieved by adding perturbation during modeling. As a result, the model is able to protect against the powerful attackers who have access to the whole dataset except the targeted sample.
This conclusion is derived from the DP's assumption that all the samples are independent, so that DP can defend the worst case scenario, i.e., except the target sample, all the remaining data ($K$-1 samples) are the prior knowledge of an attacker. 
Unfortunately, this assumption is broken in pairwise data, even if we treat a pair of data as a single sample. For example, the disease relation between Alice and Bob in Fig. \ref{fig_how_we_do} can be inferred by Jim since Jim has the same disease as both Alice and Bob. 
Therefore, attackers does not need $K$-1 samples to get the same prior knowledge as the worst case.
In the pairwise data setting, the worst case assumption ($K$-1 samples in an attacker's prior knowledge) in DP is not valid anymore.
The DP assumption, which can be seen as a defensive boundary for securing pairwise privacy, needs to be redefined to understand the defence capability of DP for  pairwise data. This thought is depicted as Fig. \ref{boundary}, where the defensive boundary would be shrunk to the target pair's boundary if data are independent.


\emph{\textbf{Our solution}.} 
This  paper  studies,  for  the  first  time,  how  pairwise information  can  be  leaked  to  attackers  under   distance  metric learning scenario. By understanding the limitation of DP's assumption on prior knowledge when preserving the privacy of pairwise data, we reformulate the worst case assumption of DP for pairwise data by introducing an extended privacy definition, namely Differential Pairwise Privacy (\DPP). A \DML algorithm satisfying \DPP can be described as Fig. \ref{fig_how_we_do}, where the input pairwise data altogether can be viewed as a graph. Compared to the traditional \DML algorithm, the novelty of the proposed \DML lies in the introduced randomness which is used to compensate for the difference (i.e., a number of edges in the graph) between the original dataset and the prior knowledge of the attacker. Therefore, the randomized \DML is able to output hardly distinguishable distance metric no matter which one is the input dataset, the original dataset or the possible prior knowledge of the attacker for the worst case. Consequently, no more useful information would be obtained for the attacker by querying the randomized \DML.

\emph{\textbf{Contributions}.} 
The key contributions of our work are as followings: 
\begin{enumerate}
    \item This paper presents \DPP, a new privacy preserving technique for pairwise data to secure distance metric learning. From the view of problem optimization, this work is shown as an important member in Empirical Risk Minimization (ERM) family. 
    \item We analyze the gradient sensitivity via exploring the distribution of pairs within a minibatch during the optimization. It helps to enhance the utility of the released distance metric by reducing the amount of injected noise.
    \item We exploit the connection of \DPP with the existing privacy research, and verify the efficacy of the proposed scheme by numerous experiments.
\end{enumerate}

The rest of this paper proceeds as follows. We present the related work in Section \ref{secRelatedWork} and introduce the preliminaries in Section \ref{secPreliminaries}. Then we state the secure metric learning problem in Section \ref{secProblemStatement}, and formally define \DPP in Section \ref{sec_Differential_Privacy_for_Pairwise_Data}. In Section\ref{sec_DML_Algorithm}, we implement a \DML algorithm based on the contrastive loss, where we further propose a utility improvement method. The experiments in Section \ref{secExperiment} are used to verify the efficacy of the proposed scheme. Some interesting questions are discussed in \ref{secDiscussion}. Finally, Section \ref{secConclusion} concludes this paper. 

\section{Related Works}\label{secRelatedWork}
We present the related works from two aspects, the possible strategies for the privacy of pairwise data, and current privacy implementation methods.

\subsection{Privacy Strategies}\label{Privacy_strategy}
Before diving into any specific privacy definitions, one may ask why not directly hiding individual identity, because it would immediately remove the privacy issues. Actually, this so-called anonymization trick has been argued as unsafe and inadequate in many previous works, such as \cite{narayanan2008robust} and \cite{sweeney2013matching}. The deficiency of anonymization is being vulnerable to the auxiliary information. For example, the information extracted from a social medium without name annotation is still possibly recognized from some other social media data with name annotation by comparison. 

A feasible choice to preserve pairwise data is to leverage Local Differential Privacy (LDP). For example, one can adopt the coin flipping style approach \cite{warner1965randomized} to probabilistically change the relationship of any pair. In this case, any subsequently surmised relationship for a data pair can be possibly denied, such as  \cite{hay2017differentially,joy2017differential,sun2018truth}. If the privacy for individual feature is needed, one can resort to the input perturbation like \cite{fukuchi2017differentially}. Unfortunately, LDP suffers the performance degeneration \cite{sun2018truth} because it fails to takes the subsequent application into consideration.

The closely related work is about studying the privacy of correlated data. One pioneer work by Kifer et al. \cite{kifer2011no} pointed out the correlated records definitely degrade the privacy level if not specially treated. They then proposed a customizable privacy framework Pufferfish \cite{kifer2012rigorous} that requires the whole algorithm should change an attacker's prior distribution as less as possible. Following this framework, Song et al. \cite{song2017pufferfish} proposed a so-called Wasserstein mechanism. Recent graph based privacy research \cite{karwa2011private,hay2009accurate,rastogi2009relationship,zhang2015private} construct the relations among individuals but mainly focus on the graphical statistics. Some other works \cite{zhu2014correlated,liu2016dependence,zhao2017dependent} in the data release community \cite{nissim2007smooth,lee2019synthesizing, zhao2019preventing} also noticed the detriment that data correlation brings in and formalized diverse DP variants. However, we emphasize that their claimed data correlation is virtually different from pairwise data. For instance, \cite{song2017pufferfish} included the time series data while \cite{zhu2014correlated} took the records distance into consideration. This means correlation in existing research is actually individual to individual. Instead, the correlation of pairwise data is caused by one individual's participation in multiple pairs. Therefore, their schemes cannot be freely extended to our problem.  



\subsection{Privacy Implementation Methods}
Given a specific privacy definition, current research suggests implementing privacy by following three ways. (1) Output perturbation. This method adds noise to the output of the algorithm, which can be easily understood when the algorithm is as simple as a database operator, like averaging the salary of the employees in a company \cite{dwork2014algorithmic}. For some machine learning models like classification or regression, \cite{chaudhuri2011differentially,zhang2017efficient} give the possible solutions. (2) Objective perturbation. By adding an extra perturbed term in the objective, \cite{chaudhuri2009privacy,chaudhuri2011differentially,kifer2012private,talwar2014private} show this strategy can works with a guarantee to the better model utility. The perturbed term is usually not general for the other objectives and needs some special considerations for different models. (3) Gradient perturbation. This line of work manage to perturb the gradient during the optimization, which has been widely used in \cite{song2013stochastic,bassily2014private,talwar2015nearly,abadi2016deep,zhang2017efficient,wang2017differentially}. This approach can be applied to most machine learning models including deep neural networks. We present the recipe of DP in machine learning via gradient perturbation. It mainly consists of 4 components. 
\begin{enumerate}
    \item Privacy principle satisfying some requirements.
    \item Loss function designed for the specific problem.
    \item Upper bound for the gradient sensitivity.
    \item Noise injection during optimization.
\end{enumerate}

\section{Preliminaries}\label{secPreliminaries}
\noindent {\bf Notation:} A pairwise datum is a tuple $z_{ij} = (\Delta x_{ij}, y_{ij})$, where $\Delta x_{ij} = x_i-x_j \, (i\neq j, \Delta x_{ij}\in \mathbb{R}^d)$ is the feature difference between the individual $i$ and $j$, and pairwise label $y_{ij}$ is a binary variable to encodes their relationship. Plus, $x_i$ could also be any representation learned by deep embedding. Omitting the subscript of $z_{ij}$ whenever it clearly indicates a single pair, we have $Z = \{z^1,z^2,...,z^K\}$ is a dataset of $K$ pairwise data which are the input of a \DML algorithm. For a vector $v$, we use $||v||$, $||v||_2$ to denote its $\ell_1$-norm and $\ell_2$-norm, respectively. In a graph, we use $\langle, \rangle$ to denote a pair of nodes and use $(,)$ to denote an edge or a path.

\theoremstyle{definition}
\begin{definition}\label{edge-disjoint}
(Edge-disjoint $s$-$t$ paths) Given an undirected graph and two nodes in it, source $s$ and destination $t$, two paths from $s$ to $t$ are said edge-disjoint if they do not share any edge.
\end{definition}
\theoremstyle{definition}
\begin{definition}\label{Lipschitz}
(Lipschitz function over model parameter $\theta$) A loss function $f:\mathcal{C}\times \mathcal{Z} \rightarrow \mathbb{R}$ ($\mathcal{C}, \mathcal{Z}$ is parameter space and input space separately.) is $h$-Lipschitz (under $\ell_1$- norm) over $\theta$, if for any $z \in \mathcal{Z}$ and $\theta_1,\theta_2 \in \mathcal{C}$, we have $|f(\theta_1,z)-f(\theta_2,z)|\le h||\theta_1-\theta_2||$.
\end{definition}

\theoremstyle{definition}
\begin{definition}\label{DPdef}
($\epsilon$-Differential Privacy \cite{dwork2006calibrating}) A randomized algorithm $\mathcal{A}$ is said to guarantee $\epsilon$-differentially private if for all datasets $\mathcal{Z}$ and $\mathcal{Z}'$ satisfying $\mathcal{D}(\mathcal{Z},\mathcal{Z}')=1$ ($\mathcal{Z}'$ is called the neighboring dataset of $\mathcal{Z}$, and $\mathcal{D}(,)$ means the number of records two datasets differ.) and for any possible algorithm output $o$ the following holds:
\begin{equation}
Pr[\mathcal{A}(\mathcal{Z})=o] \leq e^{\epsilon} \cdot Pr[\mathcal{A}(\mathcal{Z}')=o],
\end{equation}
where $Pr[\cdot]$ is w.r.t. the randomness in $\mathcal{A}$, and the non-negtive parameter $\epsilon$ is known as privacy budget.
\end{definition}

Sensitivity based methods for $\epsilon$-DP leverage the Laplace Mechanism \cite{dwork2014algorithmic} which has the following definitions. 

\theoremstyle{definition}
\begin{definition}\label{sensitivity}
($\ell_1$-sensitivity) The $\ell_1$-sensitivity of a function $g: \mathcal{Z} \rightarrow \mathbb{R}^d$ is:
\begin{equation}
\Delta g = \max_{\mathcal{D}( \mathcal{Z} , \mathcal{Z}' )=1} \| g(\mathcal{\mathcal{Z}}) - g(\mathcal{Z}') \|.
\end{equation}
\end{definition}

\theoremstyle{definition}
\begin{definition}\label{LapMechanism}
(Laplace Mechanism) Given a function $g: \mathcal{Z} \rightarrow \mathbb{R}^d$, the Laplace Mechanism $\mathcal{M}_{Lap}$ is defined as:
\begin{equation}
    \mathcal{M}_{Lap}(\mathcal{Z},g,\epsilon) = g(\mathcal{Z})+ Y,
\end{equation}
where $Y$ is drawn from Laplace distribution $Lap(0,b)$ with $b = \frac{\Delta g}{\epsilon}$. Laplace Mechanism preserves $\epsilon$-differentially private.
\end{definition}
\begin{remark}
Definition \ref{DPdef} can be relaxed to the \emph{approximate} DP according to \cite{dwork2014algorithmic} if Definitions \ref{Lipschitz}-\ref{LapMechanism} are properly modified. The proposed \DPP can be also relaxed with the same style. We leave these analyses to Appendix \ref{Appendix_sensitive} for a supplemental discussion.
\end{remark}
\noindent {\bf Distance Metric Learning.} \DML aims to learn a representative distance between $i$ and $j$ defined by $\mathcal{D}_M(i,j) = \sqrt{\Delta x_{ij}^TM\Delta x_{ij}}$, where $M \in \mathbb{R}^ {d\times d}$ is the distance metric, a.k.a., a symmetric positive semidefinite matrix. The original form of $M$ refers to the case where $i$ and $j$ are drawn from the same distribution with covariance matrix $\Sigma$, with $M = \Sigma^{-1}$. According to the literature \cite{weinberger2009distance,guillaumin2009you,ying2012distance,hu2014discriminative,sohn2016improved,xie2018nonoverlap,li2018online}, this metric can be better learned once the pairwise labels are provided. Practically, pairwise label is often denoted as a binary variable $y_{ij} \in \{0,1\}$ to indicate category. For instance, if two samples $i$ and $j$ are from the same class, then $y_{ij}=0$, and if they are from different classes, $y_{ij}=1$. Let $M = W^TW$, where the transformation matrix $W \in \Bbb{R}^{d^\prime \times d}$ ($ 1\leq d^\prime \leq d$) is free of any constraint. As a result, $W$ can be optimized by minimizing the projected distance between similar samples while maximizing the distance between dissimilar ones. \\

\section{Problem Understanding}\label{secProblemStatement}
\subsection{Insights}\label{sec_Problem_Understanding}

Since the pairwise data encodes the interpersonal relationship, preserving the privacy of involved pairwise relationship is the core problem in this work. The algorithm designer is thought as the trusted party, and thus the whole dataset can be directly fed into the algorithm, i.e., \DML. The user is returned an available distance metric after they submit a query to the system (\DML algorithm). To prevent any potential attacks in this setting, the algorithm designer is responsible to develop a \DML algorithm which guarantees the privacy of every pairwise relationship.


Practically, we are playing the role of the algorithm designer. Inspired by \cite{kifer2012rigorous}, we adopt the following principle to preserve pairwise relationship. \textbf{If an attacker fails to infer the target relationship through her/his prior knowledge, she/he cannot obtain more information by querying the \DML algorithm either.} Based on this principle, we have three key insights. 
\begin{enumerate}
    \item Defending against the attacks that need query. If a targeted sample can be inferred through its relations with other samples in the prior knowledge of the attacker, privacy preserving techniques, e.g., DP, would never be possible to defend the attacks. Therefore, as the algorithm designer, we are supposed to recognize and formulate the maximum prior knowledge of an attacker, and prevent them knowing more about data during their interactions with the \DML algorithm. In contrast, the users who can already infer the target relationship through their prior knowledge is outside of our scope, because they do not need to query a \DML algorithm output. 
    
    \item Preserving both feature difference and pairwise label. We are motivated by preserving the privacy of pairwise relationship, i.e., $y_{ij}$, the last element of the tuple $z_{ij}$, similar to \emph{Attribute DP} introduced in \cite{kifer2011no}. Unfortunately, it is actually not adequate to solely preserve $y_{ij}$. Recalling the goal of \DML, the distance change between two individuals reflects their relationship. For instance, two samples close to each other but far away in the projected space are likely labeled as dissimilar, i.e., $||\Delta x_{ij}||$ is small while $\mathcal{D}_M(i,j)$ is large. As our task is to return an available distance metric, we have to hide the information of $\Delta x_{ij}$ to avert this kind of leakage. Therefore, we conclude that to achieve the privacy of any pairwise relationship between $i$ and $j$, the privacy concern for $\Delta x_{ij}$ and $y_{ij}$ are both needed. Please note it is not equal to preserving a single tuple $z_{ij}$, and see Section \ref{PrivacyConcenonEdge} for more details. 

    \item Enhancing the utility by introducing randomness as less as possible. Existing works \cite{chaudhuri2011differentially,chaudhuri2009privacy,kifer2012private,song2013stochastic,bassily2014private,talwar2015nearly,zhang2017efficient,wang2017differentially,talwar2014private} applying DP in machine learning algorithms have shown introducing randomness to the machine learning model is provable to preserve the data privacy (Refer to the recipe in Section \ref{secPreliminaries}). Compared to DP in which randomness is only used to compensate the change of any single sample, the privacy cost for pairwise data is apparently higher because there are more than one pair's change needs compensating. As a result, the utility of output distance metric is decreased. This problem is alleviated by using a sensitivity reduction technique (See section \ref{sec_Improvement_by_Sensitivity_Reduction}) which reduces the amount of injected noise. 
\end{enumerate}

\subsection{Clarification}\label{Formalization}
\begin{figure}
 \centering
  \includegraphics[width=7.2cm]{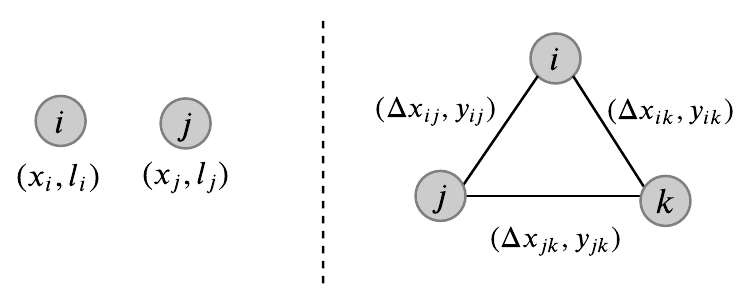}
  \caption{Comparison between the samples fed to classification or regression models and pairwise data fed to \DML algorithms. {\bf Left:} Any two samples composed of the feature $x_i$ ($x_j$) and its label $l_i$ ($l_j$) are independent in existing ERM-based works. {\bf Right:} Pairwise data are correlated with each other because an individual may participate in multiple pairs. }
  \label{figERM_COMP}
\end{figure}

We clarify two peculiar properties of \DML in this part. 

\emph{\textbf {Transitive vs intransitive relationship.}} Pairwise labels have different semantics in different context, and we notice that they are not consistent when considering their correlations. Summarily, there are mainly two types of pairwise relationship; one is transitive and the other is intransitive. Transitive relationship is like having the same disease or working in the same company, and intransitive relationship is like being friends or hanging out together. In the main body of this paper, we focus on the transitive relationship and attribute intransitive relationship as its special case. It is reasonable because transitive relationship requires more privacy concern for the transitivity risks. Please note the naive edge differential privacy (edge DP) \cite{hay2009accurate,rastogi2009relationship} is not applicable even for the intransitive relationship. We leave the analysis of intransitive relationship to Appendix~\ref{Appendix_intransitive}.

\emph{\textbf{ \DML vs classification/regression.}} A formulated \DML loss function can be seen as an instance of ERM-based \cite{vapnik1992principles} model. From this view, the transitivity property of pairwise data makes \DML distinct from existing privacy works for classification and regression \cite{chaudhuri2011differentially,song2013stochastic,abadi2016deep,wang2017differentially}. Fig. \ref{figERM_COMP} exhibits their comparison over input data. Concretely, any two data points fed to existing ERM-based models are usually independent with each other, which naturally matches the DP definition. For pairwise data, the dependence happens due to the correlation both on feature differences and pairwise labels indicated as edge on the right of Fig. \ref{figERM_COMP}. Please note although two types of correlation are caused by the same reason, they are virtually not the identical problem. The specific details are discussed in Section \ref{sec_Differential_Privacy_for_Pairwise_Data}.


\section{Differential Pairwise Privacy from Graph Perspective}\label{sec_Differential_Privacy_for_Pairwise_Data}
Denoting each individual as a node and every pairwise data $z_{ij}$ as an edge, we arrive at an \emph{undirected} graph $G = (V,E)$. Please note the node set $V$ only keeps the identity of individuals while the edge set contains all the information used for \DML training, i.e., $E = \{z_{ij} = (\Delta x_{ij},y_{ij})|(i,j \in V^2 \wedge i \neq j\}$. Thus, we can see that all the privacy burden is on edges. Based on this graph, Differential Pairwise Privacy (\DPP) is presented in this section. We will show this definition is a nontrivial extension of DP (a.k.a. edge DP in the context of graph data) over the pairwise data.

\subsection{Privacy Concern on Edge}\label{PrivacyConcenonEdge}
Let $\langle s,t\rangle$ denote the target pair whose pairwise relationship is the interest of the attacker. According to the statement in Section \ref{sec_Problem_Understanding}, given the original graph $G$, we need to formulate the prior knowledge $G'$ of the attacker.

\begin{figure}
  \centering
  \includegraphics[width=7.5cm]{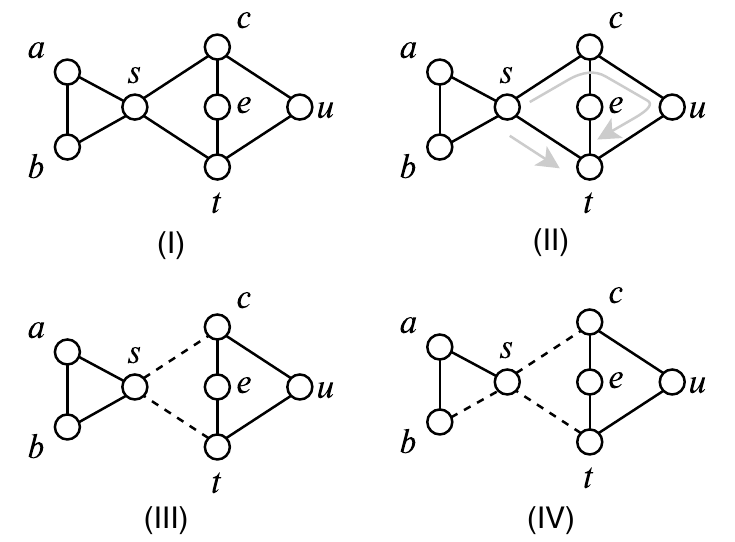}
  \caption{Construction of neighboring graph w.r.t. the pair $\langle s,t \rangle$. (\rom{1}) The graph encoding all the pairwise data. (\rom{2}) Disjoint-edge identification. (\rom{3}) Two key edges $(s,c)$ and $(s,t)$ determining the relationship inference. (\rom{4}) Edge $(b,s)$ exposing the feature of the individual $s$.}
  \label{fig3}
\end{figure}

\emph{\textbf {Pairwise relationship correlation.}} For binary category case, we summarize there are three basic relationship inference patterns\footnote{The number of inference patterns are actually determined by the category number, but even so we can still come to the consistent conclusion if similar pairs are dominant in the correlations.}. Let $\sim, \nsim$ denote that two individuals are from identical and different categories respectively, and $C$ is the category number. We have
\begin{itemize}
   \item $s \sim i,\: i \sim t \Rightarrow s \sim t$
   \item $s \sim i,\: i \nsim t \Rightarrow s \nsim t$
   \item $s \nsim i,\: i \nsim t \Rightarrow s \sim t$ (iff $C=2$)
\end{itemize}
Thus, we conclude that where there is a path between two nodes there might be a possible inference exposing their relationship. That is to say, for a target pair $\langle s,t\rangle$, all the paths between $s$ and $t$ provide useful inferences. According to Menger's Theorem \cite{goring2000short}, the least cost of preventing these inferences is properly breaking down $|\mathcal{P}_{st}|$ edges, where $\mathcal{P}_{st}$ is the set of all edge-disjoint $s$-$t$ paths. We use an example shown as Fig. \ref{fig3} to clarify this thought. Fig. \ref{fig3} (\rom{1}) shows the derived graph $G$ formed by the given pairs. Inspecting the nodes $s$ and $t$, we find there are totally three paths from $s$ to $t$, i.e., $(s,t)$, $(s,c,e,t)$, and $(s,c,u,t$). We select two of them as edge-disjoint paths which are marked as red and green line respectively in the Fig. \ref{fig3} (\rom{2}). Obviously, an instant way to prevent the inference is to break the key edges, i.e., $(s,c)$ and $(s,t)$, shown as the dashed edges in the Fig. \ref{fig3} (\rom{3}).

\emph{\textbf {Feature difference correlation.}}
According to linear algebra, when an individual is included in a circle, its feature can be calculated since the degree of freedom equals to the number of constraints (i.e., edges). As shown in Fig. \ref{fig3} (\rom{4}), the feature of $s$ is exposed once the information of edges $(s,a)$, $(a,b)$ and $(b,s)$ is known. As hiding the feature of either $s$ or $t$ is adequate to privatize $\Delta x_{st}$, one of the possible options is to delete the edge $(b,s)$ in this example. It is observed that breaking down the edge-disjoint paths sometimes help decrease the cycles used for feature inference, e.g., $s$ is also in the cycle $(s,c,e,t,s)$. Thus, the edges determining $\Delta x_{st}$ inference are only searched in the subgraph $G-\mathcal{P}_{st}$, i.e., removing all edges belonging to $\mathcal{P}_{st}$ from $G$. Let $c_s$, $c_t$ denote the number of edges that isolate the nodes $s$ and $t$ from the possible cycles over $G-\mathcal{P}_{st}$, respectively. The minimum number of edges we should delete is $\min\{c_s,c_t\}$.

Summarily, it is concluded that the attacker who targets the relationship between $s$ and $t$ should at least miss  $|\mathcal{P}_{st}|+ \min(c_s,c_t)$ edges in $G$. This quantitative measure actually generalizes the attacker's prior knowledge but makes it easy to do formulation. As any pair is the potential target pair, the attacker's prior knowledge $G'$ should satisfy
\begin{equation}\label{def_kappa}
    \mathcal{D}(G,G') \ge \kappa,
\end{equation}
where $\mathcal{D}(,)$ means the number of edges two graph differs, and the introduced variable $\kappa$ is calculated by 
\begin{equation}\label{eq:kappa}
    \kappa = \max_{\forall s,t \in V, s\neq t}{\{|\mathcal{P}_{st}|+ \min(c_s,c_t)\}}.
\end{equation}
Particularly, we name $G'$ as $\kappa$-neighboring graph of $G$ if the equality exactly holds in Eq. \eqref{def_kappa}.


\subsection{Differential Pairwise Privacy}
Given a graph $G$, we name the attacker as $\kappa$-\emph{Att} if her/his prior knowledge is exactly $\kappa$-neighboring graph of $G$.
An attacker with fewer edges prior, i.e., $\mathcal{D}(G,G') > \kappa$, is unable to know more than $\kappa$-\emph{Att}, while an attacker knowing more edges, i.e., $\mathcal{D}(G,G') < \kappa$, is likely to have known the target pair without any need of querying the \DML algorithm. Thus, $\kappa$-\emph{Att} defines the worst case for pairwise data.

From Eq. \eqref{eq:kappa}, $\kappa$ is only determined by the given dataset. Although searching the exact value of $\kappa$ is not an interactive component of the \DML algorithm, it is very challenging to compute in reality. To overcome this dilemma we propose an alternative approach to effectively calculate the value of $\kappa$. The details are left to Appendix \ref{appendix:a} for the interested readers.


Suppose $\kappa$ is known, based on the above analyses, we now give the definition of differential pairwise privacy.
\begin{definition}\label{DPPdef}
($\epsilon$-Differential Pairwise Privacy (\DPP)). A randomized \DML algorithm $\mathcal{A}_{DML}$ is said to guarantee $\epsilon$-differentially pairwise privacy if for all datasets $G,G'$ satisfying $\mathcal{D}(G,G')= \kappa$ and for any possible output $o_M$ the following holds:
\begin{equation}
Pr[\mathcal{A}_{DML}(G)=o_M] \leq e^{\epsilon} \cdot Pr[\mathcal{A}_{DML}(G')=o_M].
\end{equation}
\end{definition}
\DPP takes good advantages of DP for pairwise data. More importantly, we present its two valuable peculiarities.
\begin{itemize}
    \item Bearing individual participation. Our definition potentially allows attacker to have prior knowledge about the participation of the target individuals. We de facto privatize the participation of their pairwise relationship, i.e., the connected edge. This is consistent with common sense. For instance, if an attacker wants to know whether Alice and Bob have the same disease, he/she must know they are contained in the target dataset.
    \item Defending implicit attacks. No assumption is made to the given pairwise labeled dataset. In practice, the data are often collected and labelled by some domain experts who may be agnostic to the involved privacy issues. As algorithm designers are the trusted party, it is their responsibility to consider the possible pairwise data leakage risks. We have shown the proposed \DPP is a feasible definition to deal with the risks.

    
\end{itemize}

\section{\DPP-\DML Algorithm Design}\label{sec_DML_Algorithm}
According to the recipe of DP in Section \ref{secPreliminaries}, since a new privacy definition for pairwise data is built up, we now apply it to a specific \DML algorithm. In this section, we make contrastive loss \cite{chopra2005learning,guillaumin2009you,xie2018nonoverlap} a case study to present the details about how to design a \DML algorithm that exactly satisfies Definition \ref{DPPdef}. Then we further improve the utility of the \DML algorithm while keeping its intact privacy by proposing the sensitivity reduction approach.

\subsection{Case Study with Contrastive Loss}

Contrastive loss is a classic \DML model which measures pairwise data in a projected space for better fitting the pairwise labels. For simplicity, $z_{ij} = (\Delta x_{ij}, y_{ij})$ is denoted by $z = (\Delta x, y)$, and the loss function is written as
\begin{equation}\label{eqobjective}
{L(W,z)=} \frac{1}{2} (1-y)D_W^2 + \frac{1}{2} y \{\max(0,m-D_W)\}^2,
\end{equation}
where  $W$ is the transformation matrix, $D_W =  \|W \Delta x\|_2$ is the distance of data point $i, j$ in the projected space, and $m > 0$ is a margin threshold used for avoiding collapsed solutions. 

We follow the gradient perturbation approach applied in \cite{song2013stochastic,bassily2014private,abadi2016deep} to implement the privacy of \DML algorithm. The consideration behind this approach is that gradient computation is the unique component of \DML algorithm that interacts with the input data. 

As each row of $W$ is independent, the gradient w.r.t $W_r (r = 1,2,..., d^{\prime})$ is 
\begin{eqnarray}\label{eq_gradient}
g_r(z) =\begin{cases}W_r\Delta x \Delta x^T & y = 0\\\frac{D_W-m}{D_W}W_r\Delta x \Delta x^T & y = 1, D_W < m \\ \mathbf{0} & y = 1, D_W \ge m\end{cases}.
\end{eqnarray}
Before further computing the gradient sensitivity, we need the following lemma.
\begin{lemma} \label{lemma1}
Given a pair $(\Delta x,y)$, the gradient function $\|g_r(\cdot)\|$ is infinite if only if $y = 0$ and $\|W_r\|$ is unbounded.
\end{lemma}
\begin{proof} 
If $y=0$ and $\|W_r\| \rightarrow \infty$, clearly $\|g_r(\cdot)\|$ will be infinite.

\hspace{2em}If $y = 1$ and $D_W \ge m$, $\|g_r(\cdot)\| = 0$.

\hspace{2em}If $y = 1$ and $D_W < m$, we have
\begin{equation}
   \left.\begin{aligned}
   \|g_r(\cdot)\| & = |1-\frac{m}{\|W\Delta x\|_2}| \cdot \|W_r\Delta x \Delta x^T\|\\
   & \overset{\circled{1}}{\le}  \frac{m \|W_r\Delta x \Delta x^T\|}{ \frac{1}{\sqrt{d'}}\|W\Delta x\|} - \|W_r\Delta x \Delta x^T\| \\
   & \overset{\circled{2}}{\le} \frac{m\sqrt{d'} |W_r\Delta x|\cdot \|\Delta x^T\|}{\|W\Delta x\|}\\
   & \overset{\circled{3}}{\le} 2m\sqrt{d'}.
   \end{aligned}\right. 
\end{equation}
where $\circled{1}$ follows the fact that for any vector $u \in \mathbb{R}^{d'}$ $\|u\|_1 \le \sqrt{d'}\|u\|_2$. $\circled{2}$ utilizes the facts that every induced norm is submultiplicative, i.e., $\|W_r \Delta x \Delta x^T\| \le \|W_r \Delta x\| \cdot ||\Delta x^T||$, and second term is non-negative. \circled{3} follows the fact $\|W_r\Delta x\| \le \|W\Delta x\|$ and triangle inequality, i.e., $\|\Delta x\| \le \|x_i-x_j\| \le  2\|x_j\| \le 2 $ where $x_i$ is $\ell_1$ normalized. 

Summarizing above results, we complete the proof. 
\end{proof}

From Lemma~\ref{lemma1} we can see $g_r(\cdot)$ is bounded if $\|W_r\|$ is smaller than a const. A tender option is to impose some restriction on $W$. For example, Jin et al. \cite{jin2009regularized} employed a regularizer to measure the complexity of $W$. Yet, it fails to bound the gradient during the entire optimization. As a result, the sensitivity of gradient $g_r(\cdot)$ will be infinite according to Definition~\ref{sensitivity}. Chaudhuri et al. \cite{chaudhuri2011differentially} averted this difficulty by constraining the objective to be $h$-Lipschitz, which is simply achieved by the gradient clipping technique \cite{abadi2016deep}. We follow their schemes and formulate a new theorem. 

\begin{theorem}\label{theorem1}
If the objective function in Eq. \eqref{eqobjective} is $h$-Lipschitz w.r.t. $W_r$, the $\ell_1$ gradient sensitivity $\Delta g_r$ on a minibatch $\mathcal{B}$ is at most $\frac{2\kappa h}{|\mathcal{B}|}$, where $\kappa$ is specified by Eq. \eqref{def_kappa}.
\end{theorem}
Theorem~\ref{theorem1} is a direct extension to the Corollary 8 of \cite{chaudhuri2011differentially}.
\begin{remark}
With this gradient sensitivity, we can subsequently make \DML algorithm satisfy Definition \ref{DPPdef} by adding the noise sampled from the distribution $Lap(0,\frac{2\kappa h}{|\mathcal{B}|\epsilon})$ to the iterative batch gradient. Particularly, according to \cite{song2013stochastic}, the whole \DML algorithm will satisfy $\epsilon$ pure \DPP if every batch is disjoint.
\end{remark}
\subsection{Improvement by Sensitivity Reduction}\label{sec_Improvement_by_Sensitivity_Reduction} 
\begin{figure}
  \includegraphics[width=\linewidth]{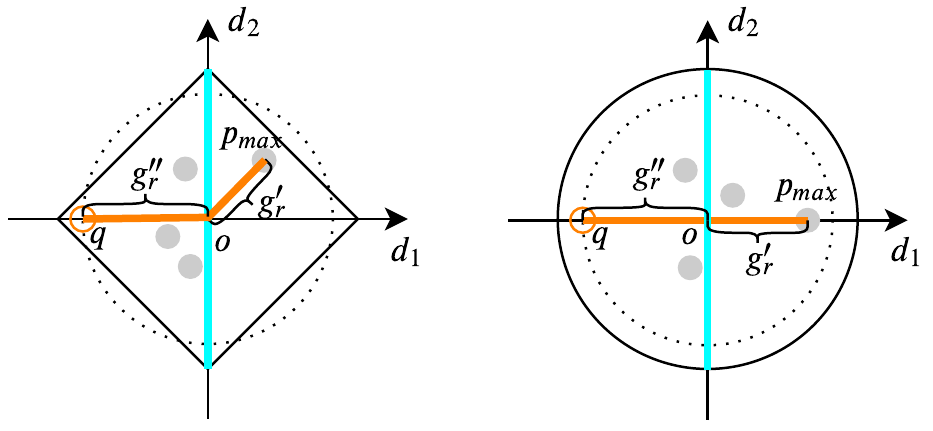}
  \caption{Gradient sensitivity reduction w.r.t. minibatch data. 
  {\bf Left:} Individuals are $\ell_1$ normalized. The cyan line segment denotes the factor $2h$ specified by Theorem~\ref{theorem1}. $p_{max}$ is the one of all batch members whose gradient value is the largest, $q$ denotes the possible counterpart of $p_{max}$ in the neighboring batch that satisfies Eq. \eqref{counterpartL1}. The orange line segment connecting $p_{max}$ and $q$ is likely shorter than the cyan one.  {\bf Right:} Individuals are  $\ell_2$ normalized, and representation are consistent with the left. This subfigure is also specified by Corollary \ref{shrink_L2} in Appendix \ref{Appendix_sensitive}.  }
  \label{figshrink}
\end{figure}

The amount of injected noise to gradient is determined by the gradient sensitivity. To improve the utility of optimization algorithm, an instant option is to reduce the gradient sensitivity value. Keeping this thought in mind we propose a gradient sensitivity reduction approach by exploring the distribution of pairs within a minibatch during the optimization.

Since we locally add noise to every batch gradient, it does not cost much to adjust the sensitivity in a minibatch. Specifically, the sensitivity based method aims to find the maximum difference between the given batch and its any possible neighbor over the gradient value. If $\kappa=1$, our goal is equivalent to find out which pair is the most sensitive among a group of batch members. We present an example in a two-dimensional space shown as Fig. \ref{figshrink}, where $p_{max}$ denotes the pair (regardless of its label dimension) who has the maximum gradient value. Without loss of generality, we suppose the larger distance between two pairs, the more different their gradients will be. Thus, Theorem~\ref{theorem1} actually employs the biggest sensitivity value represented by cyan segment. In the presented case, $p_{max}$ is the most sensitive pair because it has the largest possible distance to the $\ell_1$ norm boundary compared to other pairs. Pair $q$ represents the farthest pair possibly occurring in the neighboring batch. That means the sensitivity value is determined by the distance of $p_{max}$ and $q$. Inspired by this observation, we propose gradient sensitivity reduction approach. In the followings, instead of tuple $z$ we abuse $p$ as a pair within a batch for the input of gradient function.

\begin{algorithm}[tb]
\caption{\DPP-\DML Algorithm}
\label{alg:algorithm}
\begin{flushleft}
\textbf{Input}: Dataset $Z$ containing $K$ tuples\\
\textbf{Parameter}: Reduced dimension $d^{\prime}$, margin threshold $m$, Lipschitz constant $h$, the distance $\kappa$, batch size $|\mathcal{B}|$, privacy budget $\epsilon$, epoch number $T_{\max}$\\
\textbf{Output}: Distance metric $M$
\end{flushleft}
\begin{algorithmic}[1] 
\STATE Initialize transformation matrix $W$ randomly, step size $\eta = 1$, counter $\tau = 0$, batch index $it_{bat} = \frac{K}{|\mathcal{B}|}$, privacy budget for an epoch $\epsilon\prime = \frac{\epsilon}{T_{\max}}$
\FOR{$T = 1,2,...,T_{\max}$}
\FOR{$it = 1,2...,it_{bat}$}
\STATE{$\tau \leftarrow \tau+1$ }
\STATE{$\eta \leftarrow \frac{\eta}{\sqrt{\tau}}$}
\FOR{$r = 1,2,...,d^\prime$}
\STATE{Compute gradient $g_{r}(\cdot)$ w.r.t each $p_j \in \mathcal{B}_{it}$ by Eq. \eqref{eq_gradient}} and then do gradient clipping $\overline {g}_{r}(p_j) = g_r(p_j)/ \max(1, \frac{\|g_{r}(p_j)\|}{h})$\\
\STATE{Compute $g_r' = \max(\overline g_r(p_1),\overline g_{r}(p_2),...,\overline g_r(p_{|\mathcal{B}|}))$}
\STATE{Compute $g_r'' = \min\{h,\max(4||W_r||,2m\sqrt{d'})\} $}
\STATE{Add noise $g_r^* = \frac{1}{|\mathcal{B}|} \sum_j\overline{g}_r(p_j) +Lap(\frac{\kappa\Delta g_r}{\epsilon'}) $}, where $\Delta g_r = \frac{g_r'+g_r''}{|\mathcal{B}|}$\\
\STATE{$W_r \leftarrow W_r - \eta g_r^*$}
\ENDFOR
\ENDFOR
\ENDFOR
\STATE{$M = W^T*W$}
\end{algorithmic}
\end{algorithm}

\begin{figure*}[t]
     \centering
     \begin{subfigure}[b]{0.23\textwidth}
         \centering
         \includegraphics[width=\textwidth]{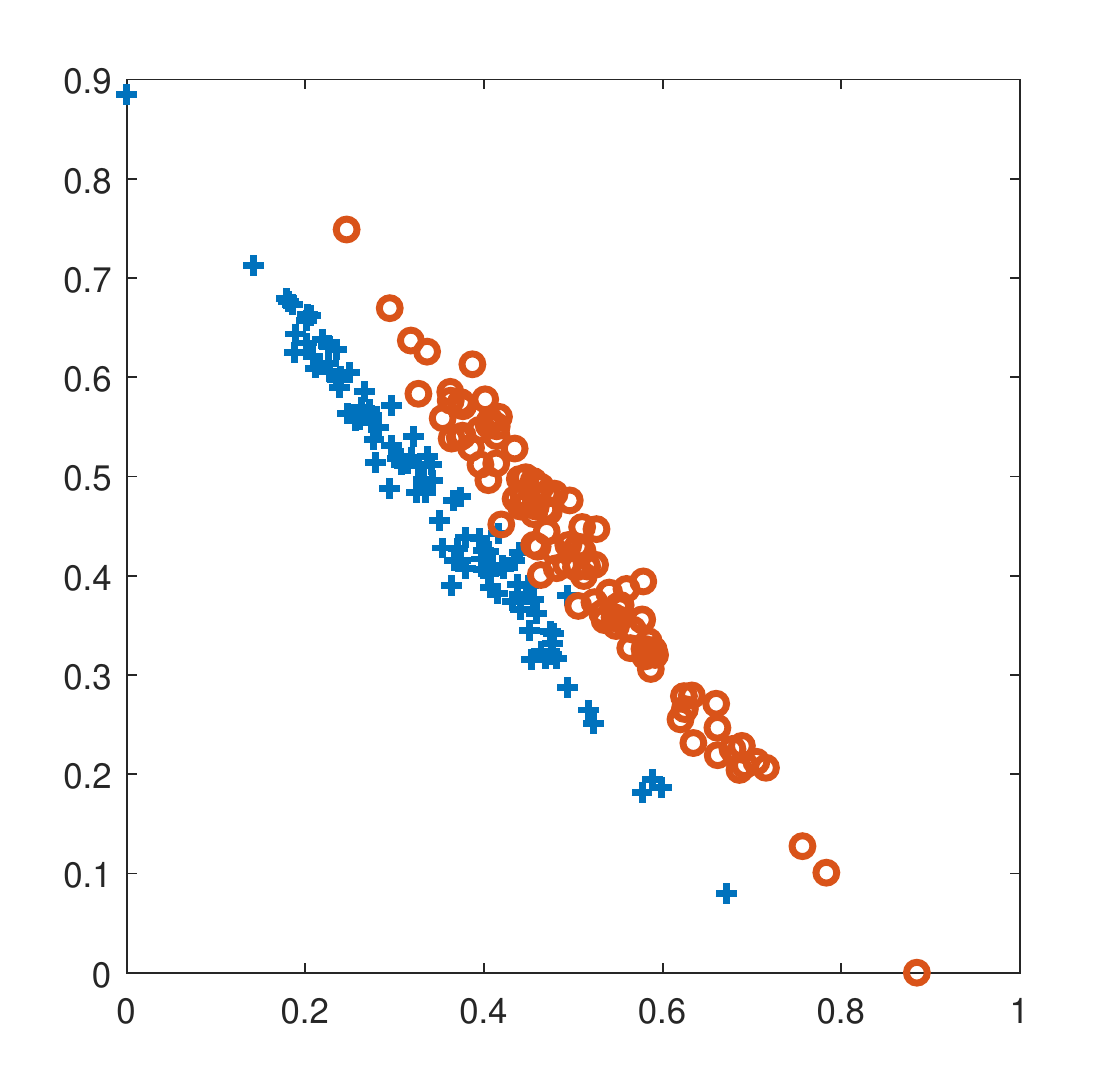}
         \caption{Original}
         \label{rawdata}
     \end{subfigure}
     \hfill
      \begin{subfigure}[b]{0.23\textwidth}
         \centering
         \includegraphics[width=\textwidth]{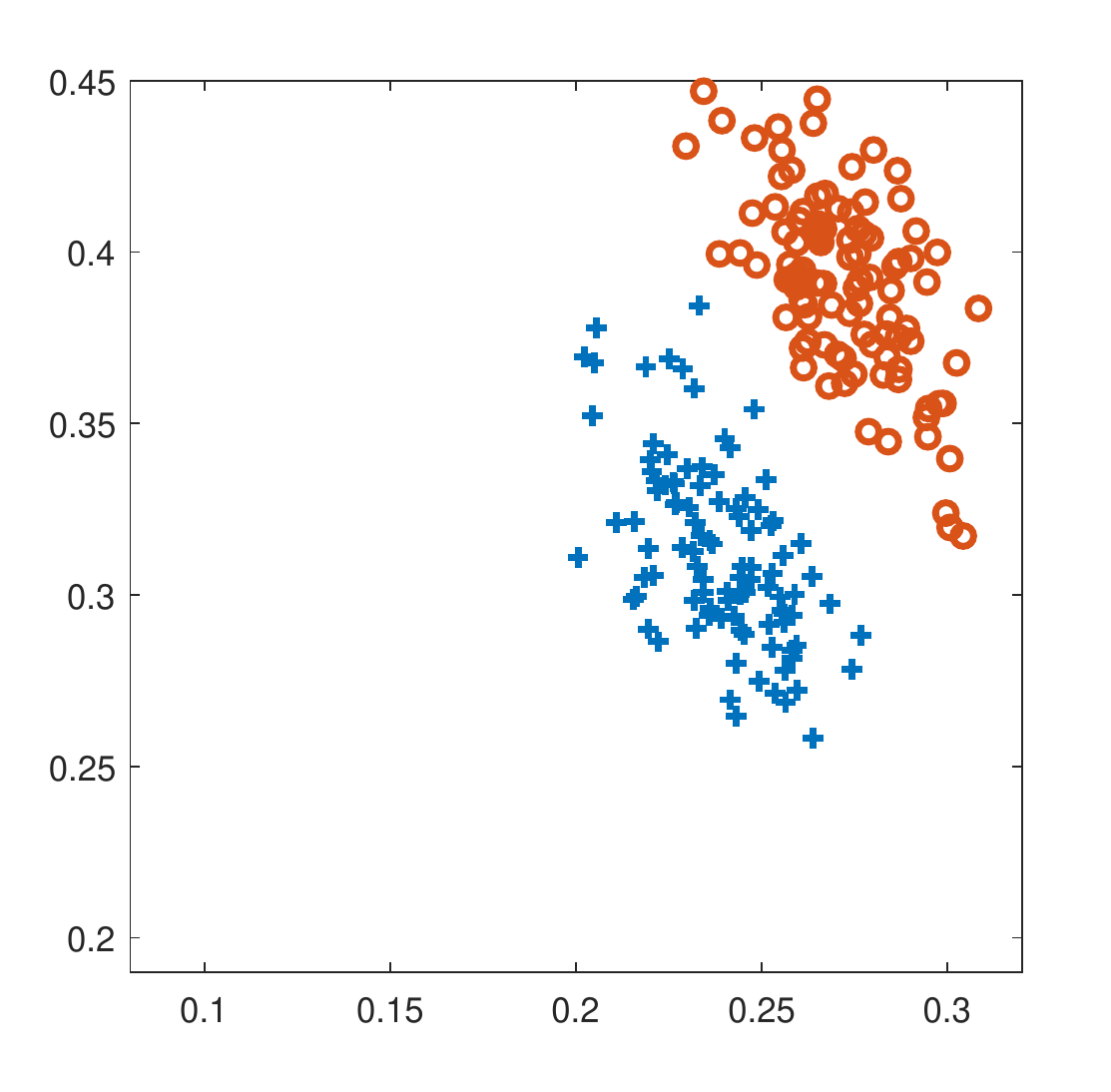}
         \caption{\DPP}
         \label{subfig-DPP-DML}
     \end{subfigure} 
     \hfill
      \begin{subfigure}[b]{0.23\textwidth}
         \centering
         \includegraphics[width=\textwidth]{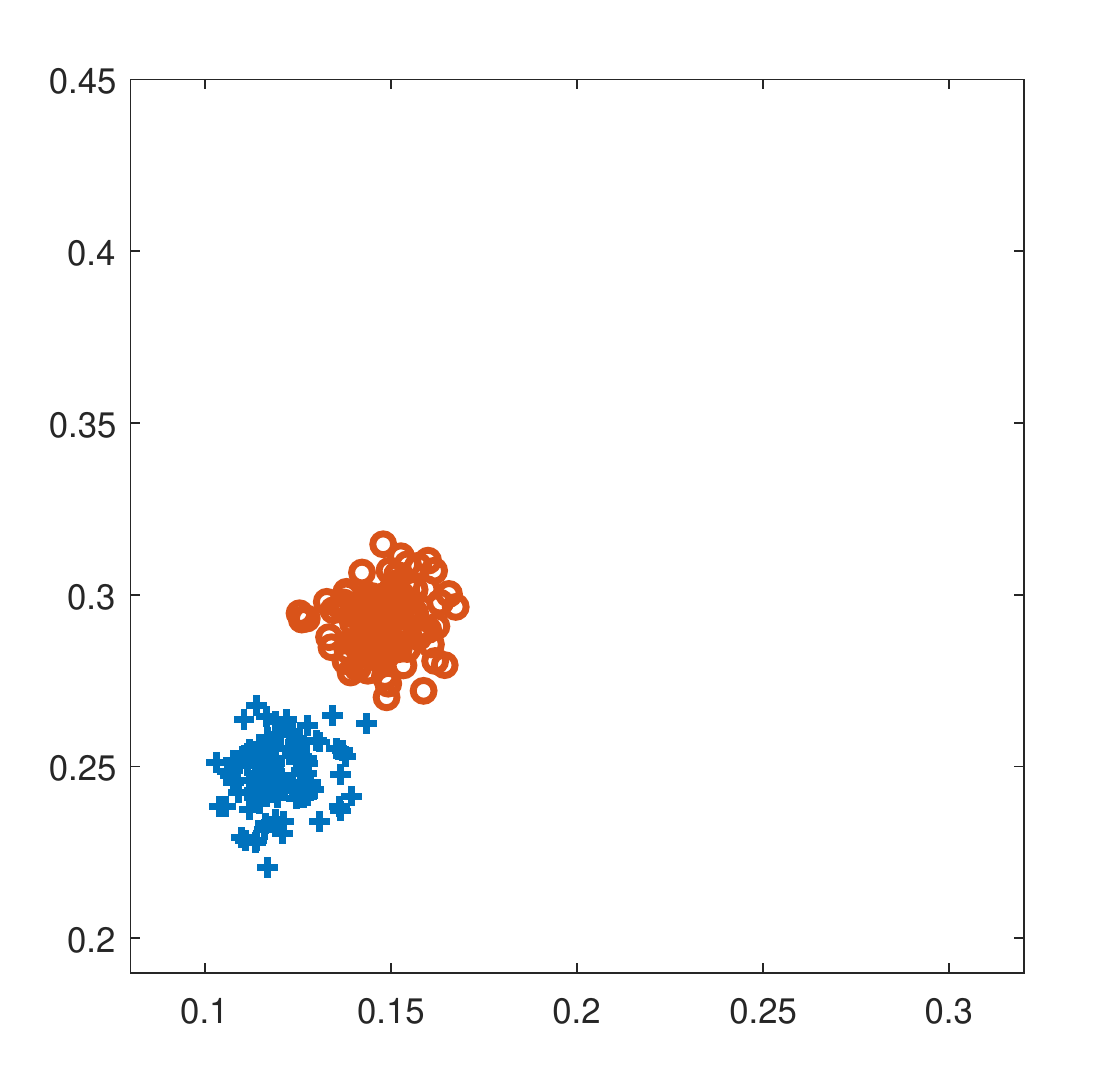}
         \caption{\DPP-S}
         \label{subfig-DPP-DML-S}
     \end{subfigure} 
     \hfill
      \begin{subfigure}[b]{0.233\textwidth}
         \centering
         \includegraphics[width=\textwidth]{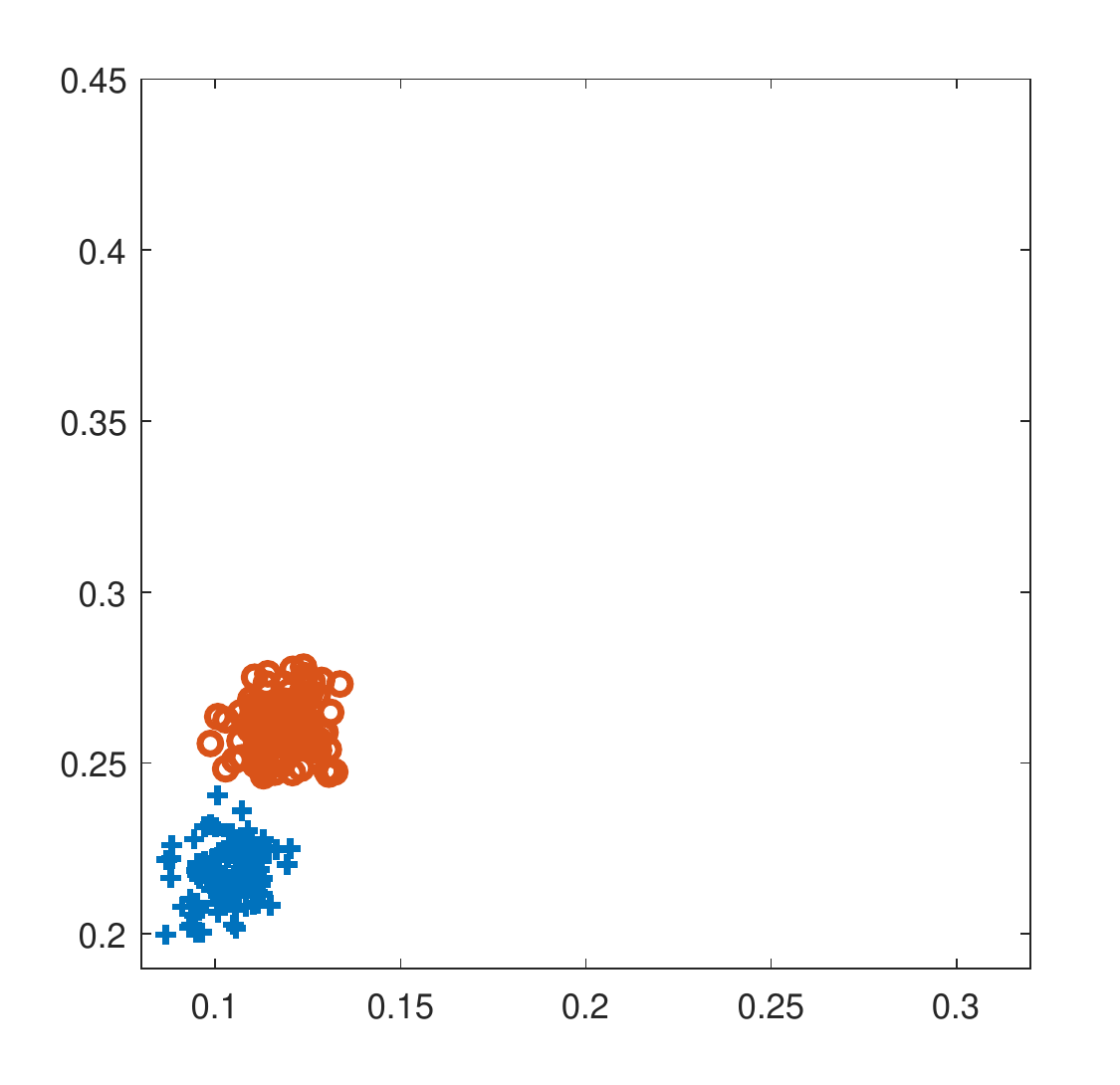}
         \caption{NonPriv}
         \label{npresult}
     \end{subfigure} 
     \caption{\DML projects original data into a new space. (a) A synthetic dataset containing 200 data points drawn from two aligned strips. (b)-(d) Data distribution after applying the metric learned by contrastive loss with \DPP, \DPP-S (with sensitivity reduction), and NonPriv concern, respectively. }\label{figEx1}
\end{figure*}

\begin{theorem} \label{theorem_shrink}
If the objective function in Eq. \eqref{eqobjective} is $h$-Lipschitz w.r.t. $W_r$, the $\ell_1$ gradient sensitivity $\Delta g_r$ on any batch $\mathcal{B}$ is at most $\frac{\kappa(g_r' +g_r'')}{|\mathcal{B}|}$, where $\kappa$ is specified by Eq. \eqref{def_kappa}, the batch gradient peak $g_r^\prime = \max(||g_r(p_1)||,...,||g_r(p_{| \mathcal{B}|})||)$, and its possible counterpart $g_r'' = \min\{h,\max(4||W_r||,2m\sqrt{d'})\}$.
\end{theorem}
\begin{proof} 
The worst case is that two neighbouring are exactly different by the constraint $ { \| \mathcal{B} - \mathcal{B}^\prime \| = \kappa}$. Since the pair satisfying $y =1$ and  $D_W \ge m$ has no contribution to the batch gradient, $\ell_1$ sensitivity of $g_r(\cdot)$ can be written as
\begin{equation}
\left.\begin{aligned}
 \Delta g_r & = \max ||\frac{1}{|\mathcal{B}|} (g_r(\mathcal{B}) - g_r(\mathcal{B}'))|| \\
 & \le \max_j \frac{\kappa}{|\mathcal{B}|} || (g_r(p_j) - g_r({p_j}'))|| \;(j=1,2,...,|\mathcal{B}|) \\
 & \le \frac{\kappa}{|\mathcal{B}|} (|| (g_r(p_{max})|| + || g_r(q))||).
 \end{aligned}\right.
\end{equation}
Let $g_r^\prime \coloneqq \max(||g_r(p_1)||,...,||g_r(p_{| \mathcal{B}|})||)$, and we have
\begin{equation}
    ||g_r(p_{max})|| = g_r' \le h.
\end{equation}
Meanwhile, according to the proof of Lemma \ref{lemma1} we have  $||\frac{D_W-m}{D_W}W_r\Delta x \Delta x^T ||\le 2m\sqrt{d'}$. As $||W_r\Delta x\Delta x^T|| \le 4||W_r||$, the following inequality must always hold
\begin{equation}\label{counterpartL1}
    ||g_r(q)|| \le \min\{h,\max(4||W_r||,2m\sqrt{d'})\} \coloneqq  g_r''.
\end{equation}
Therefore, we arrive at
\begin{equation}
\Delta g_r \le \frac{\kappa(g_r' +g_r'')}{|\mathcal{B}|},
\end{equation}
which completes the proof. 
\end{proof}
\begin{remark}
Clearly we have $g_r'= h$ if full batch gradient descent is applied. Otherwise, the proposed approach takes advantage of the difference between $g_r'$ and $h$. Please note a minibatch is usually collected from a component of $G$ ($G$ could be \emph{disconnected}) according to \cite{hu2014discriminative,sohn2016improved}, instead of globally sampling over the entire dataset $D$. Thus, $g_r'$ is diverse across different batches. It is noted that sensitivity reduction trick also benefits from the difference between $g_r''$ and $h$. The reason is that the gradient function of contrastive loss is piecewise different. We attribute this point to the  property of \DML loss function. 

\end{remark}

Combining with the steps of optimizing the contrastive loss, we summarize the entire process into Algorithm~\ref{alg:algorithm} which naturally satisfies Definition~\ref{DPPdef}. For Laplacian mechanism, it is known that $Var(||g_r||) \propto (\frac{\Delta g_r}{\epsilon'})^2$, where $\epsilon'$ is the fixed privacy budget for each minibatch. This means the smaller sensitivity factually implies the better utility of gradient. From the composition theory \cite{mcsherry2009privacy}, the privacy budget for an epoch is still $\epsilon'$. Particularly, once $T_{max}$ epochs are needed for better convergence, then the accumulated privacy budget should be $\epsilon = \epsilon'T_{\max} $. 

\section{Experiment}\label{secExperiment}
This section contains four parts. In the first part, the efficacy of the proposed method is validated on a synthetic dataset. In the second part, based on the same privacy mechanism, i.e., Laplace mechanism, we further compare with other two methods on four real-world benchmarks. Their performance is evaluated by classification accuracy on the test set. By replacing the based privacy mechanisms we demonstrate the effectiveness of sensitivity reduction in the third part. In the last part, we investigate the effects of different parameters. Through all the experiments, the $\ell_1$-norm of every sample is preprocessed smaller than 1 before used to compute the feature difference. To guarantee the convergence, all the stochastic algorithms follow the step size update rule in Algorithm \ref{alg:algorithm}, which typically enforces the conditions $\sum_\tau \eta(\tau)=\infty$ and $\sum_\tau \eta^2(\tau) \le \infty$.
\subsection{Toy Example}
We use the synthetic dataset introduced in \cite{nguyen2017supervised} as a toy example for \DML training. As shown in Fig. \ref{rawdata}, raw data are composed of two classes, each of which is single-Gaussian distributed and contains 100 points. By trickily selecting 50 intra-class pairs within each
class and 50 inter-class pairs, we obtain an undirected acylic graph. In this experiment, contrastive loss is optimized with NonPriv, \DPP, and \DPP-S (\DPP with sensitivity reduction) concern separately. To verify their capacities, in this toy example, we suppose the original data is accessible.

During the optimization, we empirically set $\epsilon = 2$, $m = 1$, $h = 0.5$, $|\mathcal{B}| = 30$, and $T_{\max} = 10$. Fig. \ref{subfig-DPP-DML}-\ref{npresult} show the transformed data using the distance metrics learned by three privacy schemes, respectively. For the convenient comparison, the transformed data are exhibited in the same range map, $[0.08,0.32]$ horizontally, and $[0.19,0.45]$ vertically. Apparently, each of them now has the clearer structure than original data. It is noticed that optimization with \DPP-S is very close to NonPriv result. That means the sensitivity reduction technique improves fidelity of output distance metric. 

\begin{figure}
     \centering
     \begin{subfigure}[b]{0.23\textwidth}
         \centering
         \includegraphics[width=\textwidth]{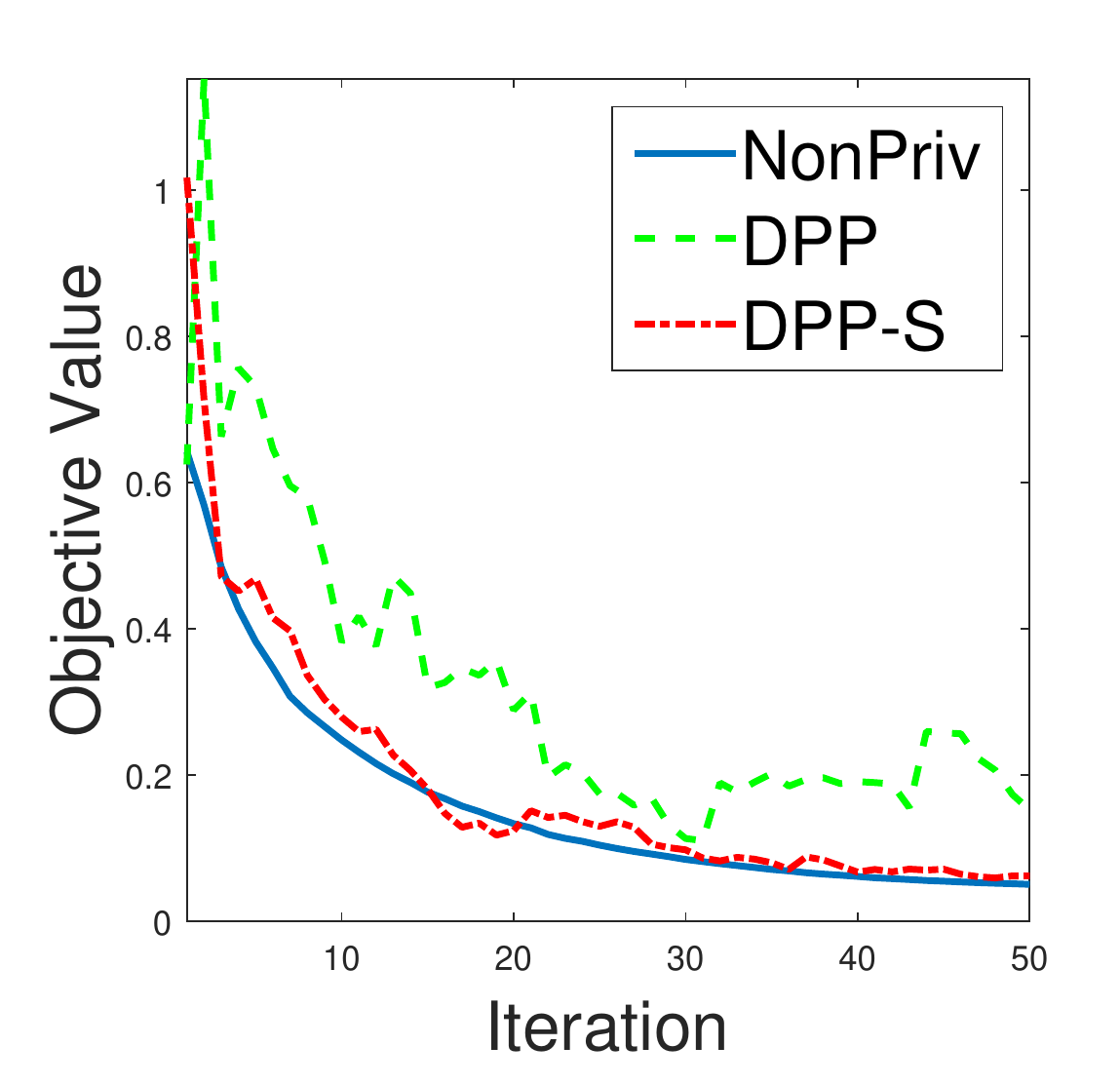}
         \caption{Convergence comparison}
         \label{toyobj}
     \end{subfigure}
     \hfill
     \begin{subfigure}[b]{0.23\textwidth}
         \centering
         \includegraphics[width=\textwidth]{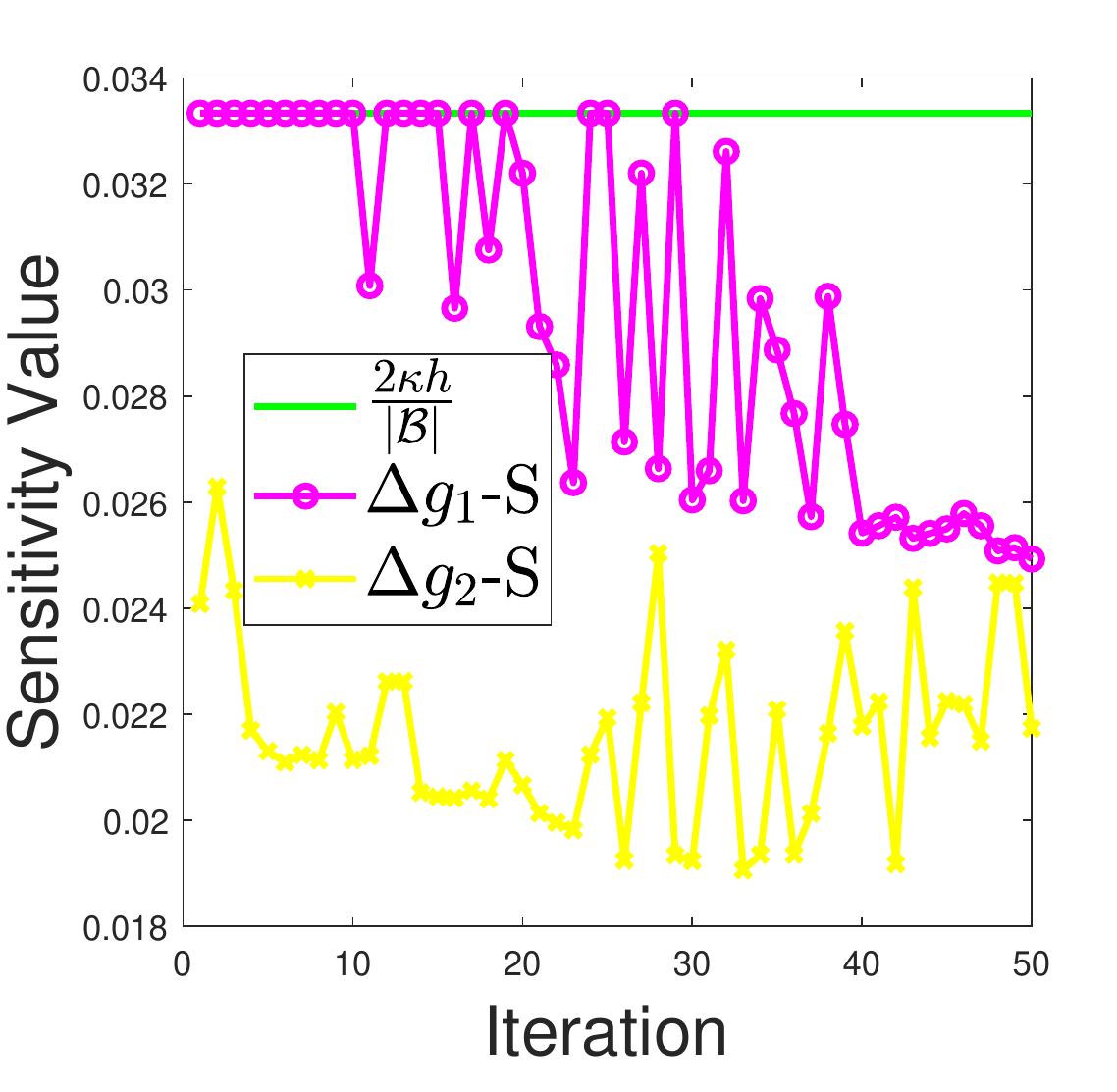}
         \caption{Sensitivity comparison}
         \label{toygra}
     \end{subfigure}
     \caption{(a) The objective values of Eq. \eqref{eqobjective} versus iteration number with NonPriv, \DPP and \DPP-S, respectively. (b) The sensitivity value $\frac{2\kappa h}{|\mathcal{B}|}$ specified by Theorem \ref{theorem1} and reduced sensitivity specified by Theorem \ref{theorem_shrink} (exhibited by each dimension) versus iteration number.  }  \label{figEx2}
\end{figure}

\begin{figure*}
     \centering
     \begin{subfigure}[b]{0.245\textwidth}
         \centering
         \includegraphics[width=\textwidth]{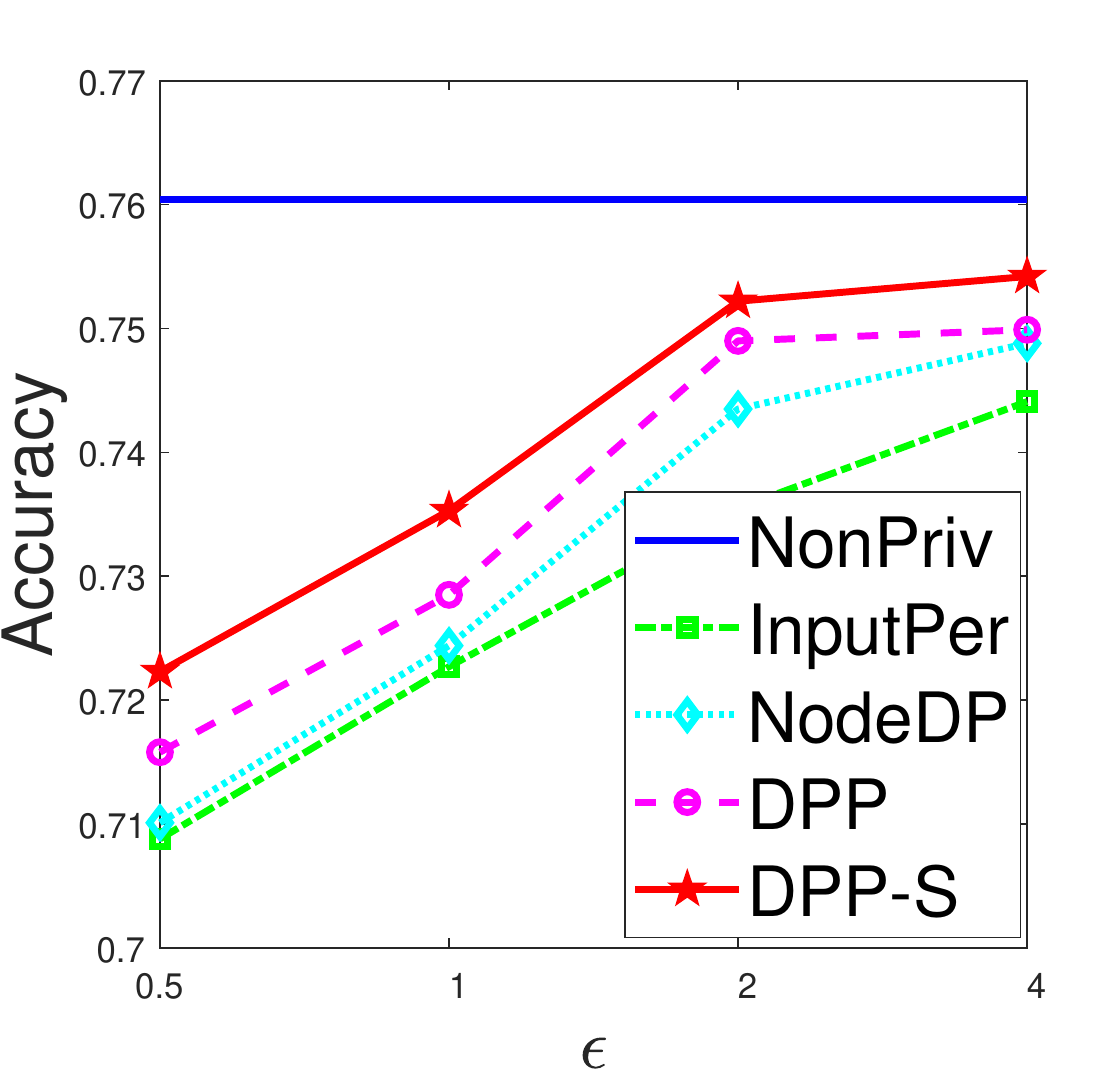}
         \caption{Adult}
         \label{acc_adult}
     \end{subfigure}
     \hfill
     \begin{subfigure}[b]{0.245\textwidth}
         \centering
         \includegraphics[width=\textwidth]{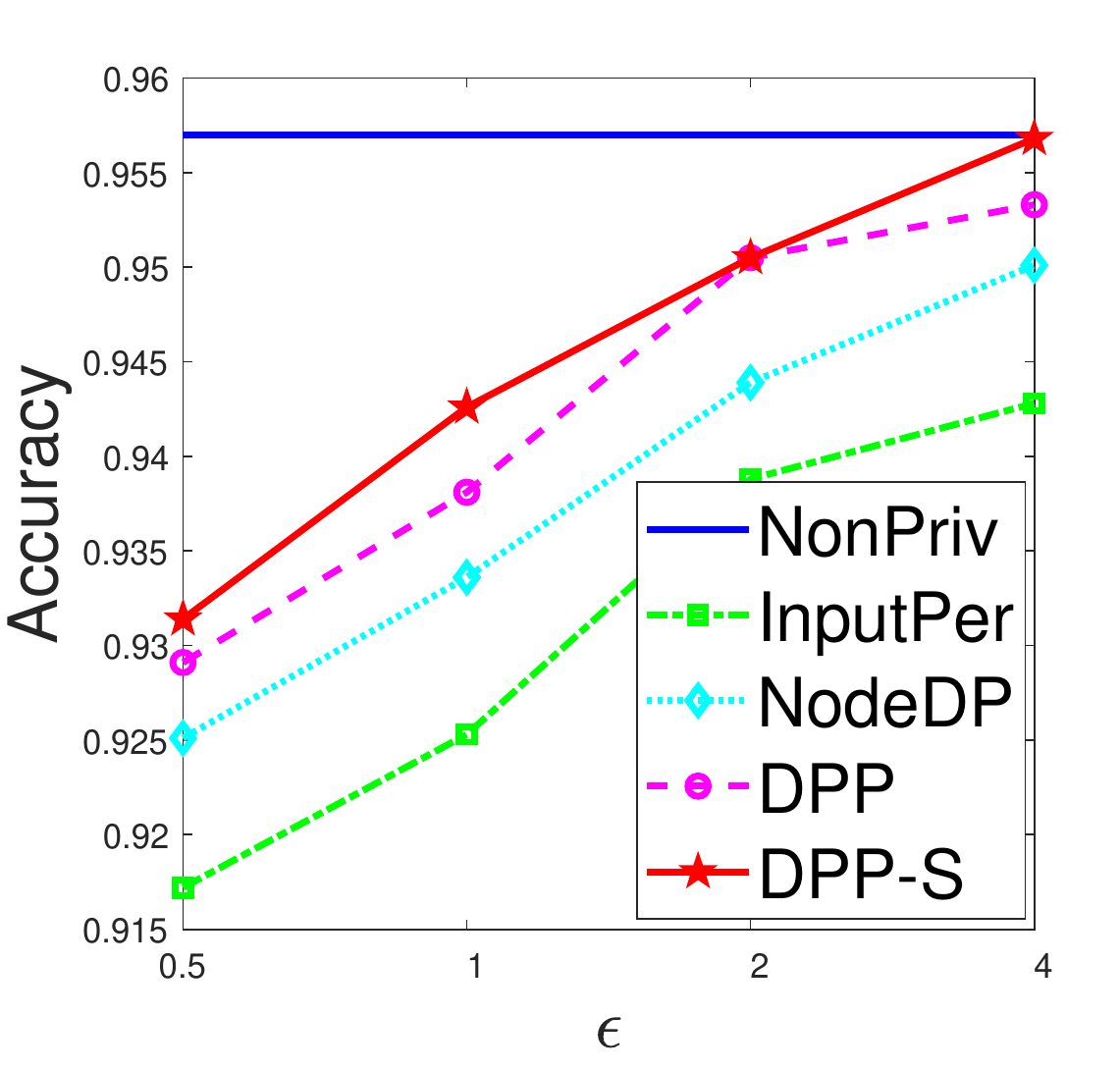}
         \caption{Bank}
        \label{acc_bank}
     \end{subfigure}
     \hfill
     \begin{subfigure}[b]{0.245\textwidth}
         \centering
         \includegraphics[width=\textwidth]{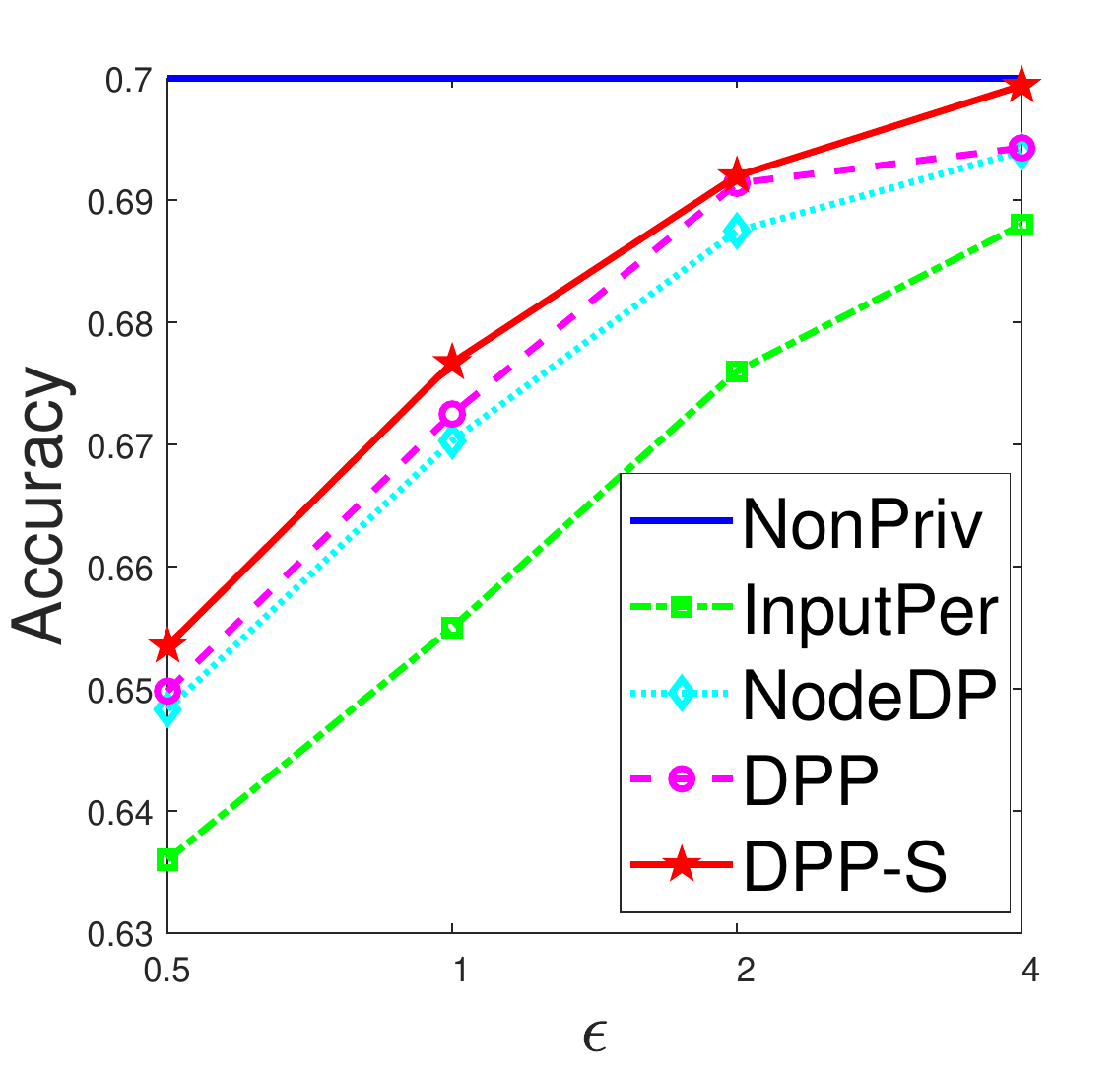}
         \caption{IPUMS-BR}
         \label{acc_br}
     \end{subfigure}
     \hfill
     \begin{subfigure}[b]{0.245\textwidth}
         \centering
         \includegraphics[width=\textwidth]{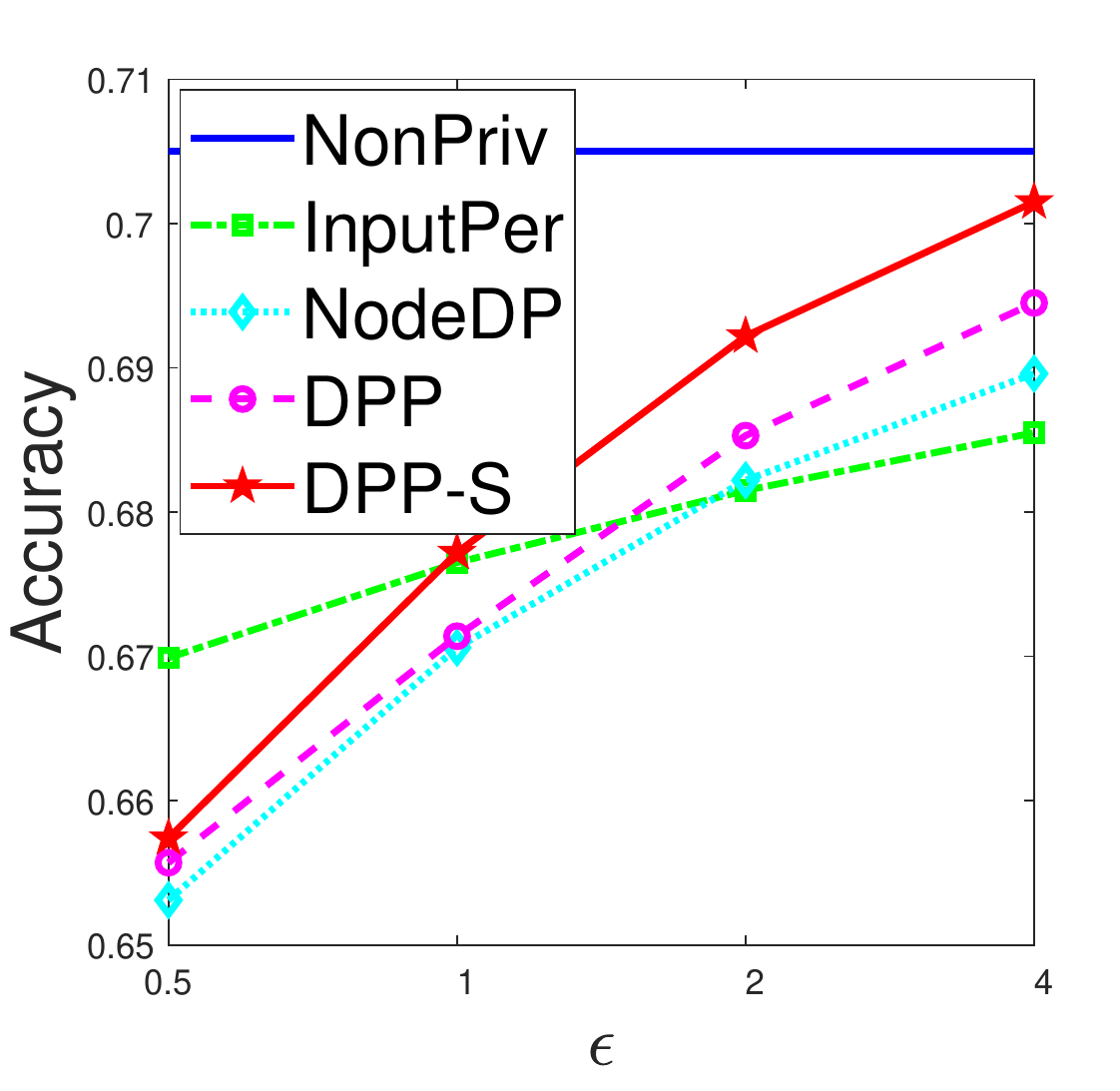}
         \caption{IPUMS-US}
         \label{acc_us}
     \end{subfigure}
        \caption{Classification accuracy of compared methods versus privacy budget $\epsilon$ over four real-world datasets.}
        \label{Real_exp}
\end{figure*}

To better understand the above observations, we present the convergence curves of each solution shown as Fig. \ref{toyobj}. It is seen that NonPriv converge steadily among three methods as its every iterative step is assigned with a clean gradient. Compared to \DPP that uses the vanilla gradient sensitivity, the objective value of \DPP-S decreases faster and has slighter fluctuation. Particularly, \DPP seems to need more rounds to converge to a steady point. It is consistent with our expectation because more noises are injected in this privacy scheme. In addition, we notice that NonPriv objective value is not strictly lower than \DPP-S. The reason is that the perturbation to the gradient sometimes provides diverse exploration directions. However, the optimization cannot benefit from random perturbations in a long period. Moreover, we investigate the sensitivity value of each row of the transformation matrix $W$, shown as Fig. \ref{toygra}. During a quantity of iterations, the sensitivity value after reduction is dramatically lower than the standard one referred in Lemma~\ref{lemma1}. We demonstrate the sensitivity reduction mitigates the utility degeneration caused by a huge amount of redundant noise.


\subsection{Comparison on Real-world Datasets}
\label{sub_rd}

Four datasets about humans are employed here following the setting of recent work \cite{lee2018concentrated}: (i) Adult \cite{Dua:2017} is extracted from the 1994 Census database which contains the instances of personal income information. (ii) Bank \cite{Dua:2017} is related with direct marketing campaigns of a Portuguese banking institution. (iii) IPUMS-BR and (iv) IPUMS-US datasets are also about Census data collected from IPUMS-International \cite{StevenR}. For imbalanced dataset (Adult and Bank), we simply downsample the majority class during training. Table \ref{table2} briefly summarizes the statistics of these datasets.  
\begin{table}[t]
\centering
\caption{Statistics of datasets}
\label{table2}
\begin{tabular}{lccc}\hline
 Dataset & Records (n) &  Dime. & Pos./Neg. Pairs \\\hline
 Adult & 48,842 & 124 & 18,700/18,700\\
 Bank &  45,211 &  33 & 8,464/8,464   \\
IPUMS-BR & 38,000 & 53 & 30,016/30,016\\
IPUMS-US & 40,000 & 58 & 31,104/31,104 \\\hline  
\end{tabular}
\end{table}

Since there is not any direct solution applying to the proposed secure metric learning problem, we borrow the idea of \emph{Node DP} \cite{kasiviswanathan2013analyzing} and ERM-based input perturbation (denoted by InputPer for short) \cite{fukuchi2017differentially} thought as two competing methods. The former defends against the attacker who is unknown as many as the maximum node degree edges, which is shown as the worst case of \DPP according to the analysis in Appendix \ref{appendix:a}. The latter assumes that data collector is also not reliable that guarantees a stronger privacy. The advantage of InputPer for pairwise data preserving is that no special consideration is needed for the pair correlations, because any inference is now unreliable. Throughout all the involved randomized algorithm, Laplace mechanism is used to extract noise. In particular, for InputPer method, the laplacian noise is added to every dimension over feature, while the label is randomly flipping by Warner's model \cite{warner1965randomized}. To prevent $\kappa$ being too large, we empirically label the pairwise data with the fixed density of constructed graph, i.e., $\frac{|E|}{|V|} = 2$. Batch size and Lipschiz const are set as $|\mathcal{B}| = 50$ and $h = 0.5$ respectively. For different datasets, the margin $m$ is preset as the average distance of dissimilar pairs, i.e., $m = \frac{1}{K_N} \sum ||\Delta x||$, where $K_N$ is the number of dissimilar pairs. The output distance metric is then evaluated by classification accuracy. Specifically, the distance metric $M$ is firstly decomposed into transformation matrix $W$, and $W$ projects original data into a new space. Then a $k$NN classifier ($k = 5$) is trained on the samples that are only employed in training pairs, and all the remaining samples are regarded as the test set. Last but not least, every method is repeated for 20 times and the average classification accuracy is eventually reported. 

Fig. \ref{Real_exp} shows the classification results comparison versus different privacy budgets. At the first glance, the classification accuracy of every randomized method rises steadily with the increment of privacy budget. It is apparent because less noise is injected to the gradient and the precision of distance metric is improved with the same iterative steps. As Node DP injects more noise than the proposed \DPP during each iteration, we find that its accuracy is always lower than \DPP. Furthermore, \DPP with sensitivity reduction, i.e. \DPP-S, effectively enhances the accuracy, especially on Adult and IPUMS-US datasets. We observe that in most cases, InputPer shows the worst performance, because InputPer is agnostic to the downstreaming application. Interestingly, it outperforms other competitors on IPUMS-US dataset when $\epsilon$ is not large. A reasonable explanation is that the dataset distance $\kappa$ or maximum node degree is sometimes large (determined by randomly labelled pairwise data) in other three methods, which results in a great amount injected noise in optimization, while InputPer is not influenced by these factors. 

\subsection{Privacy Mechanisms Comparison}
\begin{figure}
     \centering
     \begin{subfigure}[b]{0.227\textwidth}
         \centering
         \includegraphics[width=\textwidth]{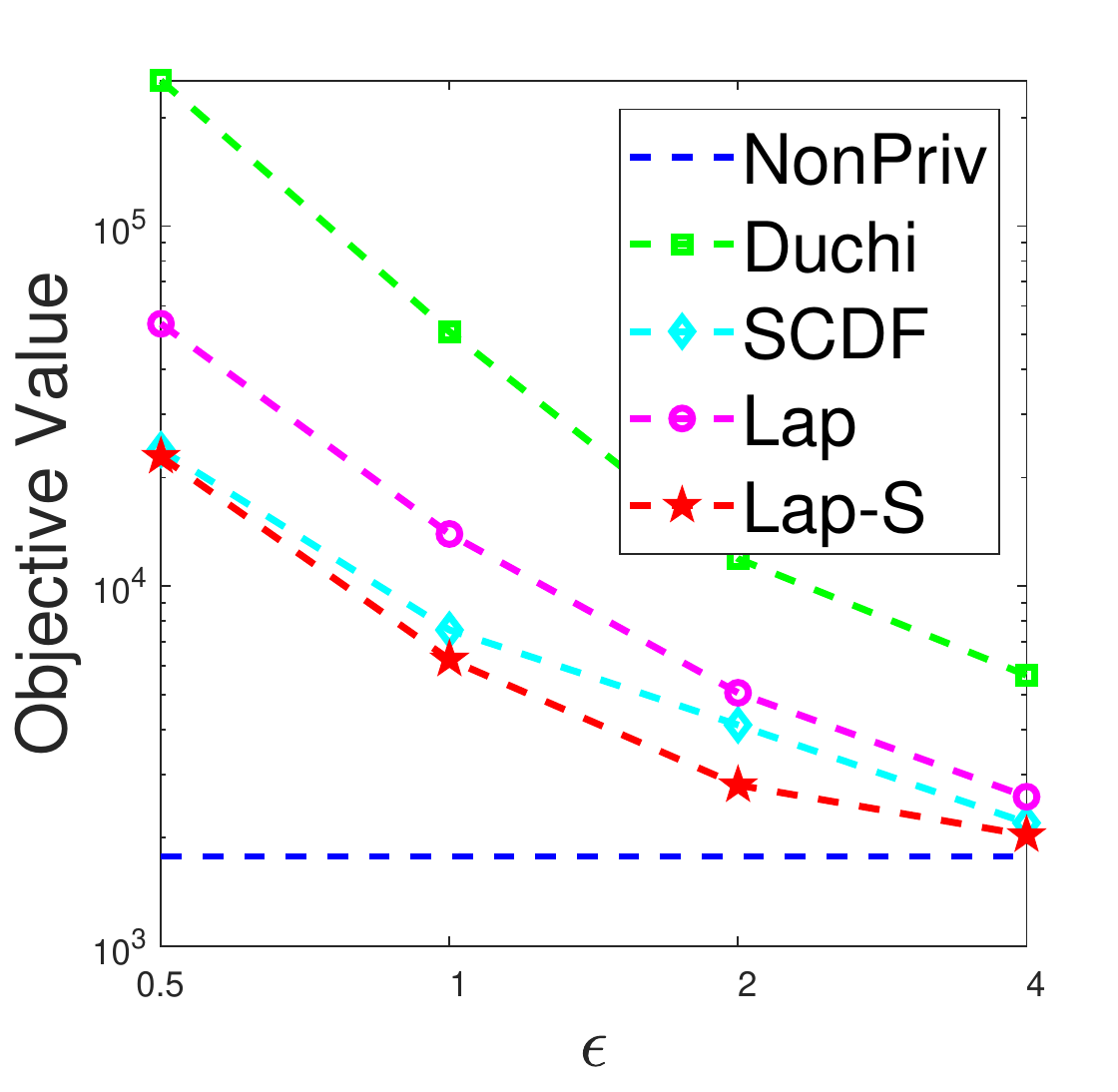}
         \caption{Adult}
         \label{para1}
     \end{subfigure}
     \hfill
     \begin{subfigure}[b]{0.23\textwidth}
         \centering
         \includegraphics[width=\textwidth]{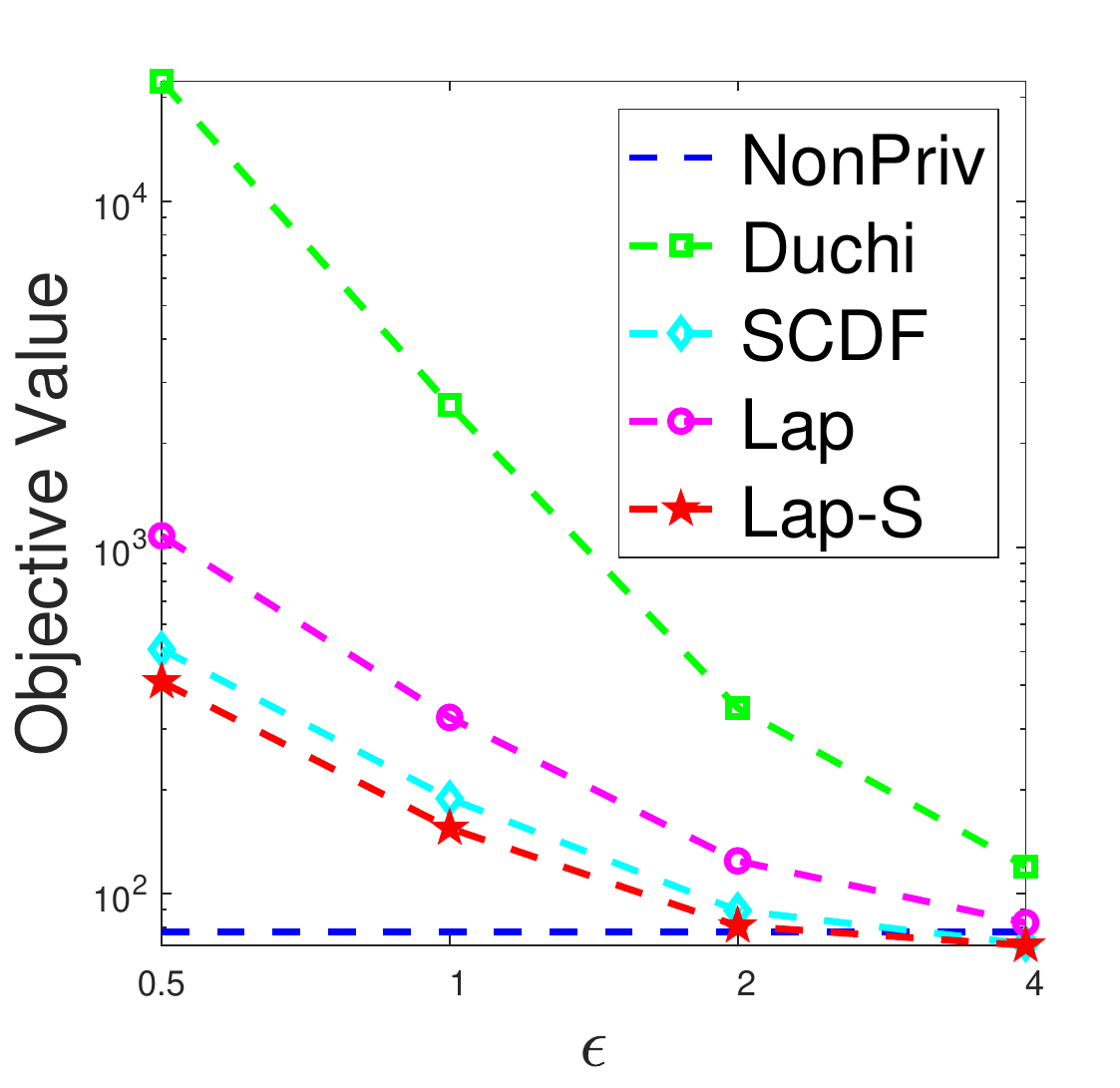}
         \caption{Bank}
         \label{para2}
     \end{subfigure}
     \caption{Different $\epsilon$-DP mechanisms comparison in implementing \DPP through their objective values.} \label{mecha}
\end{figure}
Besides Laplace mechanism (and its sensitivity reduction version), we employ two more $\epsilon$-DP mechanisms as the alternatives during implementing \DPP. Soria-Comas and Domingo-Ferr \cite{soria2013optimal} proposed a staircase-like mechanism which is a variant of Laplace mechanism. Duchi et al. \cite{duchi2018minimax} proposed a method to perturb multidimensional numeric tuples. For convenience, these four mechanisms are orderly denoted by Lap, Lap-S, SCDF, and Duchi separately. We compare them by replacing the gradient perturbation component and record their objective values after an epoch optimization. Obviously, the smaller objective values indicates better convergence. For simplicity we only do this group of experiments on Adult and Bank datasets. The involved parameters setting follows Section \ref{sub_rd}.

Fig. \ref{mecha} presents the objective values of Eq. \eqref{eqobjective} when different privacy mechanisms are employed. On both two datasets, we observe that SCDF converges faster than Lap. This is because SCDF benefits from an adjustable parameter controlling staircase width in Laplace mechanism. Staircase width is a function of both sensitivity and privacy budget. Thus, it has the potential to add less noise than Lap that enhances the utility. In addition, with the use of reduced sensitivity trick, we find that Lap-S converges fastest in all mechanism members. Duchi performs worst in this experiment compared with other sensitivity based mechanisms. Specifically, all the mechanisms are extracting unbiased noise, but the variance of Duchi for one-dimension data is  $(\frac{e^{\epsilon}+1}{e^{\epsilon}-1})^2$ while the variance is $\frac{4h^2}{|\mathcal{B}|^2 {\epsilon}^2}$ for Lap. If we take $h = 0.5$, $|\mathcal{B}| = 50$ and $\epsilon = 1$, Duchi's variance is around 10,000 times larger than Lap's!

\subsection{Effects of Parameters}
\begin{figure*}
     \centering
     \begin{subfigure}[b]{0.227\textwidth}
         \centering
         \includegraphics[width=\textwidth]{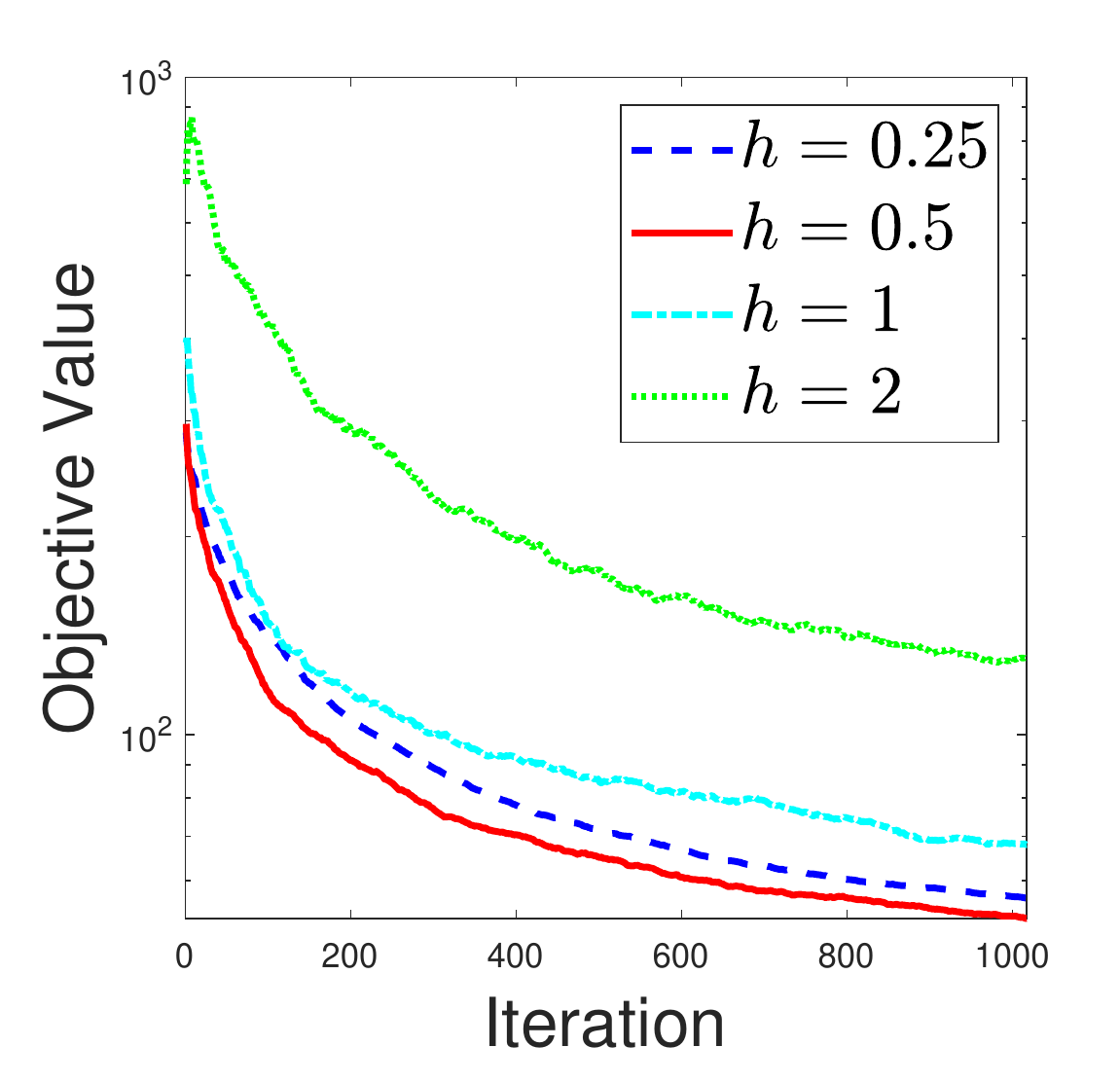}
         \caption{Lipschitz constant }
         \label{para1}
     \end{subfigure}
     \hfill
     \begin{subfigure}[b]{0.225\textwidth}
         \centering
         \includegraphics[width=\textwidth]{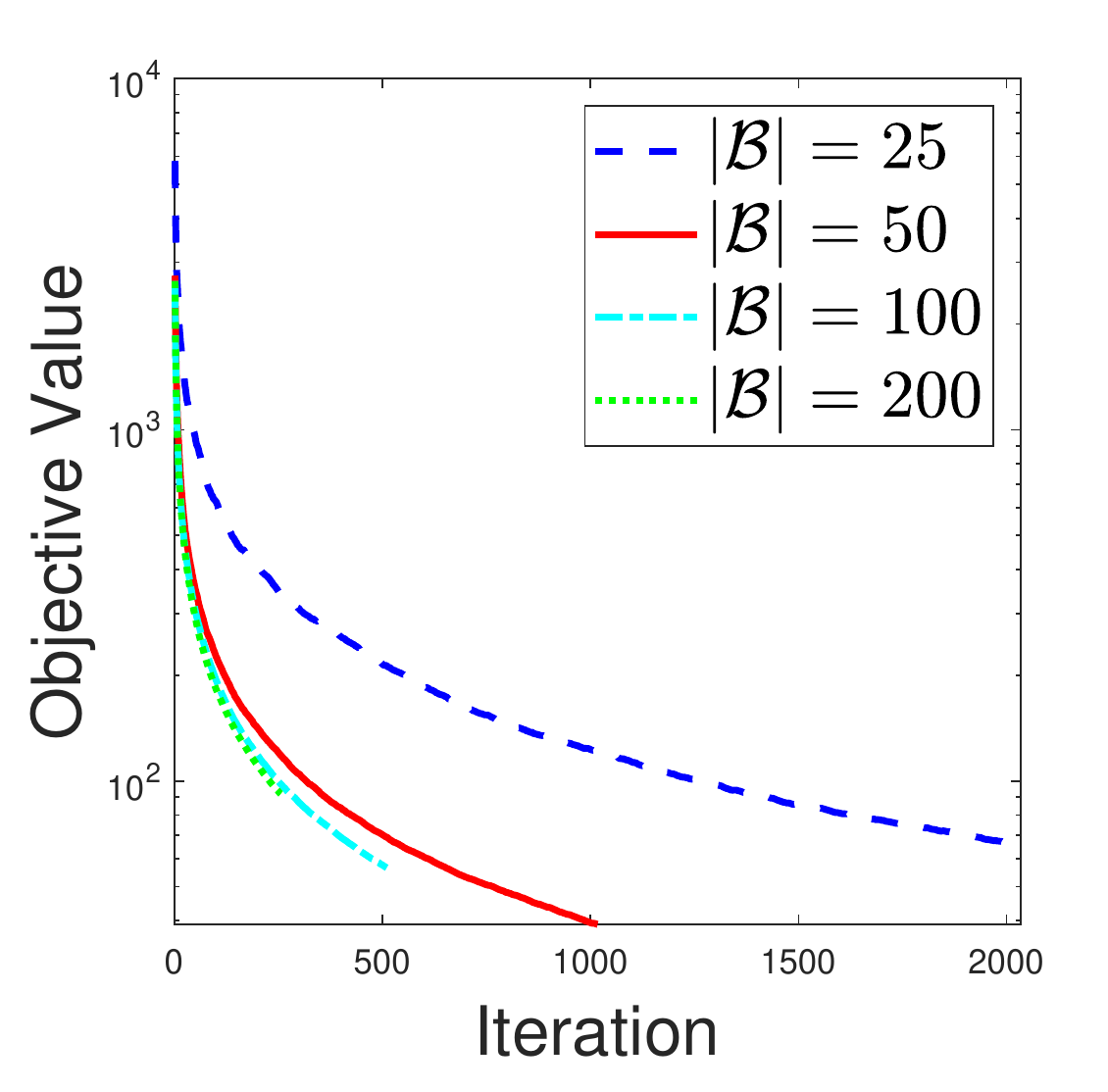}
         \caption{Batch size}
         \label{para2}
     \end{subfigure}
         \hfill
     \begin{subfigure}[b]{0.23\textwidth}
         \centering
         \includegraphics[width=\textwidth]{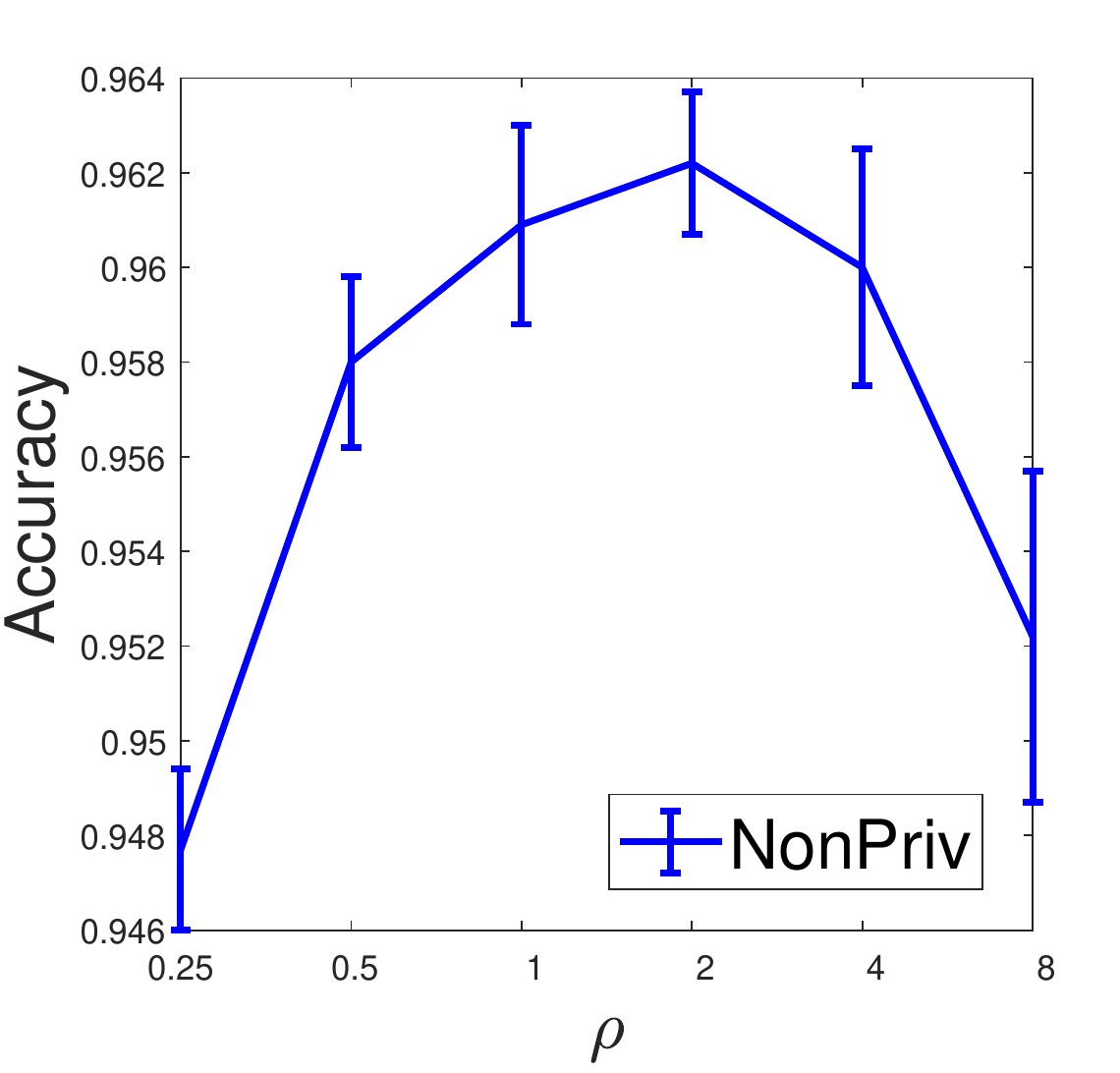}
         \caption{Margin threshold }
         \label{para3}
     \end{subfigure}
         \hfill
     \begin{subfigure}[b]{0.225\textwidth}
         \centering
         \includegraphics[width=\textwidth]{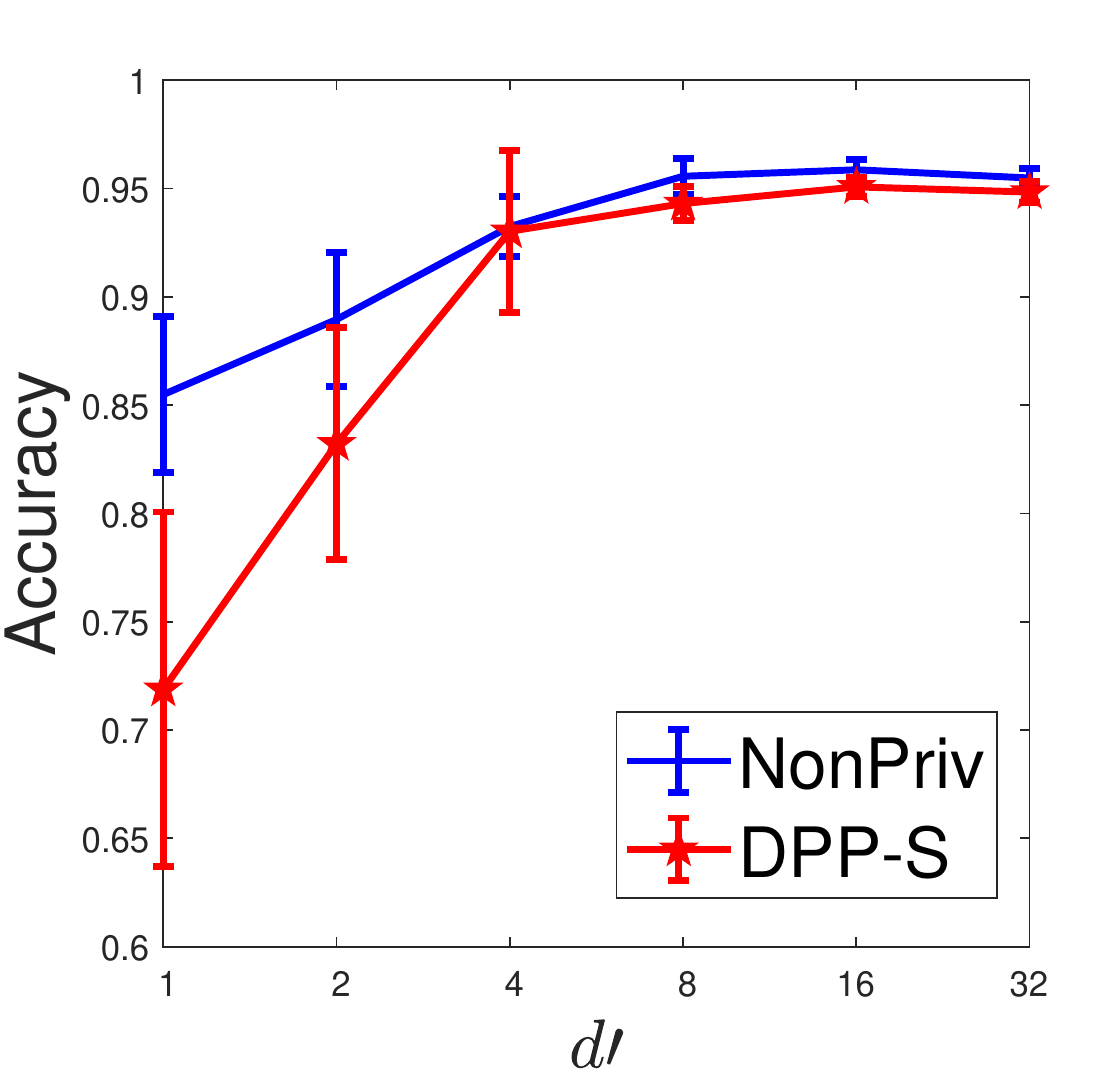}
         \caption{Dimension reduction}
         \label{para4}
     \end{subfigure}
     \caption{Effects of several key parameters on Bank dataset.} \label{para}
\end{figure*}
We simply demonstrate the effects of involved parameters Lipschitz constant $h$, batch size $|\mathcal{B}|$, margin threshold $m$, and reduced dimension $d^{\prime}$ on Bank dataset. 

To properly clip the gradient in each step, we need a good Lipschitz constant $h$, because either a smaller or larger $h$ would cause the slower convergence of optimization. On the one hand, if $h$ is too small, then most of gradients will be clipped, and thus the objective converges slowly. On the other hand, if $h$ is too large, the sensitivity value will be large according to Theorem \ref{theorem_shrink}, and consequently the amount of noise is increased in each step optimization.  In this experiment, let $\epsilon = 4$ and $T_{\max} = 3$, we investigate the objective curves under different values of $h$. The experimental results are shown as Fig. \ref{para1}. We can observe the objective converges faster when $h=0.5$. Particularly, when $h = 2$, the objective value initially increases within the a few of steps and then decreases steadily. It is because the step size is large at the beginning of the optimization, which amplifies the imprecise gradient caused by a relatively large $h$.

Similar to Lipschitz constant $h$, according to Theorems \ref{theorem1} and \ref{theorem_shrink}, the batch size is also related to the convergence rate of the algorithm. We do grid search for $|\mathcal{B}|$ and all of them are conducted for same epochs, i.e., $T_{\max} = 3$. The results are shown as Fig. \ref{para2}. When the batch size is larger, the total optimized steps are smaller, and vice versa. As $|\mathcal{B}| = 50$ reaches the lowest objective value within the limited epochs, we empirically accept it as the default setting in other experiments.

The margin threshold $m$ is a hyperparameter for contrastive loss. Instead of manually fixing it, we connect it with the average distance of dissimilar pairs, i.e., $m =  \frac{\rho}{K_N} \sum ||\Delta x||$, where $\rho$ serves as a ratio factor. In this experiment, we tune $m$ by changing the value of $\rho$ for NonPriv. Since the objective value is a function of $m$, we search $\rho$ by evaluating the corresponding testing accuracy. Fig. \ref{para3} shows the accuracy comparison. Although $\rho = 2$ obtains the best performance, the accuracy of $\rho = 1$ is only slightly lower than the best result. Thus, we simply use the naive average distance in all of other experiments.

A distance metric can be decomposed into transformation matrix, which projects the original data into different dimensional space, a.k.a. dimension reduction. Fig. \ref{para4} presents the testing accuracy result versus different reduced dimensions. The highest accuracy is obtained when $d' = 16$, and the projection without dimension reduction ($d' = 32$) is close to the best performance. Interestingly, it is observed that with the decrease of dimension, the variance of testing accuracy increases significantly. We conclude most of features are contributive in this dataset and each category data are likely discriminative when more dimensions are kept.



\section{Discussion}\label{secDiscussion}
Although we have made contrastive loss a case study in this paper, it is believed that the proposed \DPP serves as a generalized privacy definition for metric learning, such as triplet loss based works \cite{weinberger2009distance,chechik2010large}, $(C+1)$-tuplet loss \cite{oh2016deep}, and $C$-pair loss objective \cite{sohn2016improved}. However, we point out the work like \cite{xing2003distance} cannot be privatized during the optimization because it employs an iterative projection strategy (to positive/negative pairwise labeling set) that will explicitly expose the pairwise information. More importantly, if the learning model is neural network based, we can follow the recipe provided by \cite{abadi2016deep} to simply replace the corresponding gradient sensitivity over a \emph{lot} instead of a batch.

Apart from \DML, there are many other machine learning algorithms that use pairwise data during training. Similar to \DML, constrained clustering methods have the better clustering performance if a small number of pairwise labeled data is offered. The pairwise data provide \emph{cannot-link} or \emph{must-link} side information which is useful to adjust the clustering structure. Thus the proposed \DPP applies to this case. In crowd-sourcing community, workers are allowed to pairwise label their preferences. Suppose privacy concern for workers' preference is desired, we can also define the pairwise privacy for this kind of application.




\section{Conclusion}\label{secConclusion}
This is a pilot work aiming to preserve the privacy of pairwise data for distance metric learning. From the view of empirical risk minimization, this work, the first of its kind in machine learning community, targets the input data that are correlated. In the paper we first study how the pairwise data could be leaked during training a distance metric learning model, then and we propose a new privacy definition to this problem. The key idea of our method is utilizing the graph property to real the involved correlations, which helps to construct the prior knowledge of the real-world attacker. We claim this work is different from previous privacy on graphs. Particularly, we clarify the connections between the proposed privacy and two classical privacy definitions over graph, edge differential privacy and node differential privacy. Some interesting conclusions have been presented in the paper.



There are two interesting open problems that need further exploration. One is how to design a general privacy implementation method that is not limited to the exact optimization strategy. The other problem is how to use some database statistics rather than pairwise data to do distance metric learning, because it is expected have the higher utility.


%

\appendices \label{secappendices}
\section{An efficient approach for deriving the $\kappa$ in Eq. \eqref{eq:kappa}} \label{appendix:a}
According to Fig. \ref{fig3} (\rom{4}), we find that the privacy for pairwise relationship and feature difference can be both eventually attributed to the 1-hop neighbors of the target node, i.e. $s$ or $t$. Inspired by this observation, we present the following proposition.
\begin{proposition}\label{proposition1}
For a target pair $(s,t)$, concerning the correlation both on pairwise relationship and feature difference, we have
\begin{equation} \label{eq:proposition1}
    |\mathcal{P}_{st}|+ \min\{c_s,c_t\} \leq De(s)-Co_+(\bar{s}),
\end{equation}
where $De(s)$ means the degree of $s$ and $Co_+(\bar{s})$ represents the increased number of components by removing $s$ from $G$. Particularly, the equality holds if and only if $c_s = c_t$.
\end{proposition}
\begin{proof} 
If all the 1-hop neighbors of $s$ are connected to $t$, there must exist a node $t' \in V$ that causes $De(s)$ edge-disjoint $s$-$t'$ paths, i.e. $|\mathcal{P}_{st}| \le |\mathcal{P}_{st'}| = De(s)$. Otherwise, $|\mathcal{P}_{st}| < De(s)$ always holds. As a result, there are at most $De(s)-|\mathcal{P}_{st}|$ edges that can provide feature inference of $s$. From the graph theory, the increased number of components $Co_+(\bar{s})$ should satisfy the equality $Co_+(\bar{s}) = De(s)-|\mathcal{P}_{st}|- c_s $. Due to the fact that $\min{(c_s,c_t)}\le c_s$, we arrive at Eq. \eqref{eq:proposition1}.
\end{proof}

For all the possible pair, we have
\begin{equation}
    \kappa' = \max_{\forall s \in V} \{ De(s)-Co_+(\bar{s}) \},
\end{equation}
where $\kappa'$ is a upper bound of $\kappa$ in Eq. \eqref{eq:kappa}. As the component number of a graph can be efficiently calculated, the whole time complexity for searching $\kappa'$ is $\mathcal{O}(|V|(|V|+|E|))$. Furthermore, one can greedily search two connected nodes having the maximum sum for the right part of Eq. \eqref{eq:proposition1} to force $\kappa'=\kappa$. 

More importantly, from Proposition~\ref{proposition1}, we instantly conclude that \DPP is upper bounded by the known \emph{Node DP} \cite{hay2009accurate}. Deleting a node in graph equals to removing all the edges associated with this node. Thus, we have $\kappa = \max_{\forall s\in V}{De(s)}$ from the view of Node DP. We conclude that the proposed \DPP is superior to Node DP. Particularly, if the derived graph is a tree with large degree node, $\kappa$ is 1 for \DPP while $\kappa$ is the maximum node degree for Node DP.

\section{\DPP for Intransitive Relationship Case} \label{Appendix_intransitive}
The intransitive relationship is a relaxed version of transitive relationship. For the pairwise data with intransitive relationship, we only need to consider the correlation on feature difference. For a target pair $\langle s,t\rangle$, if $s$ and $t$ are mutually 1-hop neighbors, then there should be $1+\min(c_s,c_t)$ edges that need to concern for the worst case, where $c_s$ and $c_t$ is searched on the subgraph $G-\langle s,t\rangle$. Otherwise, only the number of $\min(c_s,c_t)$ edges need concerning, where $c_s$ and $c_t$ are searched over the entire graph. It is noted that the former case equals to the transitive relationship when $\mathcal{P}_{st} = \{(s,t)\}$. For the latter case, similar to Proposition \ref{proposition1} we have 
\begin{equation}
    \min(c_s,c_t) \le De(s)-Co_+(\bar{s}) -1.
\end{equation}
This inequality follows the fact that $c_s = De(s)-Co_+(\bar{s}) -1$. One can define $\kappa$-neighboring graph of $G$ now by assigning the greater value of two cases to $\kappa$. Since the value of $\kappa$ is determined by the given pairwise data, edge DP cannot handle this case.

\section{Sensitivity Shrinking For Approximate \DPP}\label{Appendix_sensitive}
We first prepare the basic ingredients for approximate \DPP. Suppose the individual features are $\ell_2$ normalized, i.e. $||x_i||_2\le 1$. We modify the $\ell_1$-norm in Definitions~\ref{Lipschitz} and \ref{sensitivity} into $\ell_2$-norm and draw noise $Y$ from Gaussian distribution $\mathcal{N}(0,\sigma^2I_d)$ with $\sigma \ge \frac{\sqrt{2\ln(1.25/\delta)}\Delta g}{\epsilon}$, which naturally updates Definition~\ref{DPdef} to be the approximate DP.

From the gradient function in Eq. \eqref{eq_gradient}, if $y = 1, D_W < m$ we have
\begin{equation}
   ||g_r(\cdot)||_2  = |1-\frac{m}{||W\Delta x||_2}| \cdot ||W_r\Delta x \Delta x^T||_2
    \le 2m.
\end{equation} 
Consequently, we have the following Corollary to calculate the sensitivity
for the approximate \DPP by extending Theorem~\ref{theorem_shrink} (also refer to the right of Fig. \ref{figshrink}).
\begin{corollary} \label{shrink_L2}
If the objective function in Eq. \eqref{eqobjective} is $h$-Lipschitz ($\ell_2$ norm) w.r.t $W_r$, the $\ell_2$ gradient sensitivity $\Delta g_r$ on any batch $\mathcal{B}$ is at most $\frac{\kappa(g_r' +g_r'')}{|\mathcal{B}|}$, where the batch gradient peak $g_r^\prime = \max(||g_r(p_1)||_2,...,||g_r(p_{| \mathcal{B}|})||_2)$, and its possible counterpart $g_r'' = \min\{h,\max(4||W_r||_2,2m)\}$.
\end{corollary}

The benefit of concerning approximate \DPP is that the total privacy cost can be reduced by slightly increasing the failure of probability of $\delta$. Since the proposed \DPP definition is consistent with DP over structure, the advanced composition theory in \cite{dwork2014algorithmic} is naturally inherited in our work. Furthermore, we refer interested readers to \cite{jayaraman2019evaluating} for a comparison of more DP variants which provide improved composition theories.



\ifCLASSOPTIONcaptionsoff
  \newpage
\fi

\bibliographystyle{IEEEtran}
\bibliography{IEEEabrv}

%








\end{document}